\documentclass[sigconf]{acmart}
\AtBeginDocument{%
  }

\setcopyright{none}
\copyrightyear{2026}
\acmYear{2026}

\usepackage{multirow}
\usepackage{comment}
\usepackage{graphicx}
\usepackage{url}
\usepackage{subcaption}
\usepackage{tcolorbox}
\usepackage{enumitem}

\begin{document}

\title{Context Selection for Hypothesis and Statistical Evidence Extraction from Full-Text Scientific Articles}

\author{Sai Koneru}
\email{saidileep543@gmail.com}
\orcid{0000-0001-8582-5914}
\affiliation{%
  \institution{Pennsylvania State University}
  \city{University Park}
  \state{Pennsylvania}
  \country{USA}
}

\author{Jian Wu}
\orcid{0000-0003-0173-4463}
\email{j1wu@odu.edu}
\affiliation{%
  \institution{Old Dominion University}
  \city{Norfolk}
  \state{Virginia}
  \country{USA}
}

\author{Sarah Rajtmajer}
\email{smr48@psu.edu}
\orcid{0000-0002-1464-0848}
\affiliation{%
  \institution{Pennsylvania State University}
  \city{University Park}
  \state{Pennsylvania}
  \country{USA}
}

\renewcommand{\shortauthors}{Koneru et al.}





\begin{abstract}
Extracting hypotheses and their supporting statistical evidence from full-text scientific articles is central to the synthesis of empirical findings, but remains difficult due to document length and the distribution of scientific arguments across sections of the paper. The work studies a sequential full-text extraction setting, where the statement of a primary finding in an article's abstract is linked to (i) a corresponding hypothesis statement in the paper body and (ii) the statistical evidence that supports or refutes that hypothesis. This formulation induces a challenging within-document retrieval setting in which many candidate paragraphs are topically related to the finding but differ in rhetorical role, creating hard negatives for retrieval and extraction. Using a two-stage retrieve-and-extract framework, we conduct a controlled study of retrieval design choices, varying context quantity, context quality (standard Retrieval Augmented Generation, reranking, and a fine-tuned retriever paired with reranking), as well as  an oracle paragraph setting to separate retrieval failures from extraction limits across four Large Language Model extractors. We find that targeted context selection consistently improves hypothesis extraction relative to full-text prompting, with gains concentrated in configurations that optimize retrieval quality and context cleanliness.
In contrast, statistical evidence extraction remains substantially harder. Even with oracle paragraphs, performance remains moderate, indicating persistent extractor limitations in handling hybrid numeric-textual statements rather than retrieval failures alone. 
\end{abstract}



\maketitle

\section{Introduction}

Hypotheses are central units of scientific method. They express testable expectations about relationships between constructs and variables \cite{wu2025research}, guiding how studies are designed, analyzed, and interpreted. In scientific articles, hypotheses are typically stated conceptually and may be paraphrased across multiple sections \cite{hesselbach2012guide}, while their empirical support is reported as statistical evidence which are hybrid numeric-textual statements with effect directions, coefficients, test statistics, confidence intervals, and p-values \cite{lang2015basic}. Extracting these hypotheses together with the evidence used to evaluate them is therefore a key step toward the scalable synthesis of empirical findings. However, remarkable growth in the number of publications each year has made it increasingly difficult to synthesize these claims at scale \cite{bornmann2021growth}, motivating automated approaches. 

Prior efforts toward this goal have primarily focused on claim extraction to support various downstream applications, including automated literature review \cite{marshall2019toward,marshallautomating}, hypothesis generation \cite{spangler2014automated}, and fact-checking \cite{wadden2020fact}. Although recent advances in Large Language Models (LLMs) have shown promise for text summarization and question-answering \cite{brown2020language}, much of existing work on scientific statement extraction has largely focused on identifying claim like sentences within abstracts \cite{achakulvisut2019claim, wei2023claimdistiller}. This abstract-level focus does not address hypothesis–evidence extraction, where hypotheses and their supporting statistical evidence often appear in different parts of a paper's full text.

\begin{figure}[h]
\centering
  \includegraphics[width=0.45\textwidth]{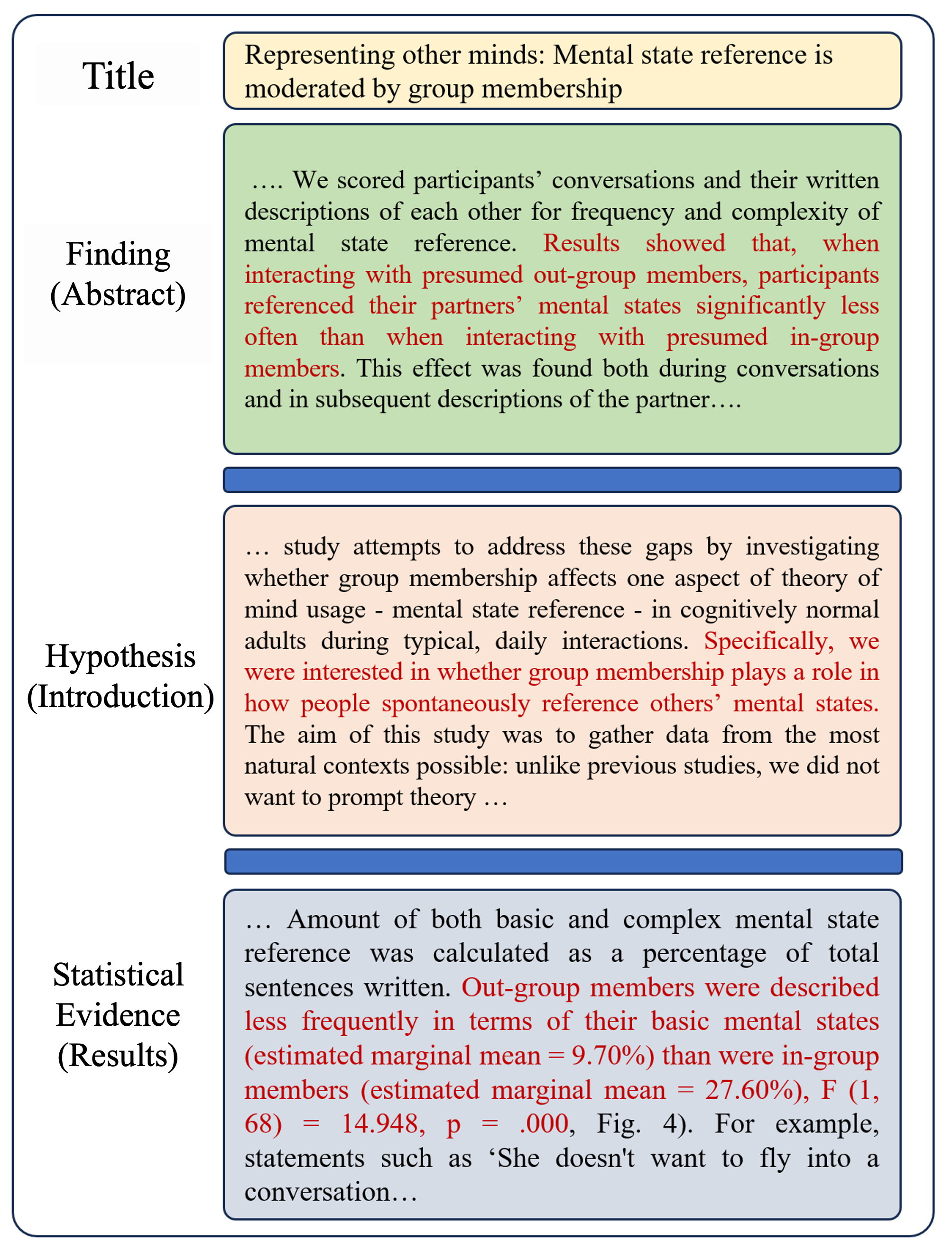}
  \caption{An example claim trace spanning the Abstract, Introduction, and Results sections of a paper. Our two-stage pipeline uses an abstract-level finding to extract a hypothesis statement (from the Introduction, in this case) and then locate statistical evidence (Results).} 
  \label{fig:experiments}
\end{figure}

Full-text extraction is challenging not only because of document length, but also because scientific reasoning is distributed across sections and embedded in layered argumentation and diverse rhetorical structures \cite{li2021scientific}. Although most state-of-the-art LLMs have sufficient context windows large enough to fit the full text of a scientific paper as input in a single pass \cite{beltagy2020longformer, xiong2023effective}, existing evaluations show that end-to-end performance is well below human level \cite{starace2025paperbench, zhang2025academiceval, cui2025curie}. In full text, the relevant hypothesis or evidence is often sparse, paraphrased, and surrounded by topically related but non-target text, so accurate extraction depends on careful context selection.

%
Our focus in this work follows the structure of data introduced by \citet{alipourfard2021systematizing}, which links a high-level \emph{abstract finding} to (i) a corresponding \emph{core hypothesis} in the paper body, and (ii) the \emph{statistical evidence} that supports that hypothesis. 
Specifically, we structure claim trace extraction into two consecutive sub-tasks that differ in their characteristics. \emph{Hypothesis extraction} is a semantic alignment task; hypotheses may be paraphrased, expressed at varying levels of specificity, and distributed across multiple sections. In contrast, \emph{evidence extraction} focuses on identifying spans containing dense statistical information (e.g., numerical values, test results), which typically appear in Results (Figure~\ref{fig:experiments}). 

We propose a consistent two-stage retrieve-and-extract pipeline (Figure~\ref{fig:two_stage_schematic}). In Stage 1, we use the pre-extracted finding statement in the abstract as a query to retrieve candidate paragraphs, from which an LLM extracts a single core hypothesis or its paraphrased variants corresponding to the finding statement. In Stage 2, we construct a composite query by concatenating the abstract finding with the extracted hypothesis to retrieve candidate paragraphs for statistical evidence extraction. We position this two stage pipeline within the Retrieval Augmented Generation (RAG) paradigm at each stage and compare it against full-text prompting baselines to isolate the effects of context selection.

Our key results are obtained through extensive controlled ablations within this paradigm, varying: 
(i) \emph{retrieval quantity}, retrieving top-$k$ paragraphs ($k \in \{5,10,20\}$); (ii) \emph{retrieval quality}, comparing standard dense retrieval, reranked retrieval, and a fine-tuned retriever paired with the same reranker; and (iii) \emph{extractor limitations} via stage-specific oracle contexts that provide the gold hypothesis paragraph in Stage 1 and the gold evidence paragraph in Stage 2, bounding performance and separating retrieval failures from extraction failures.  We find: 
\begin{itemize}[leftmargin=*]
    \item \textbf{Task-dependent bottlenecks.} Hypothesis extraction is strongly bounded by retrieval quality and context cleanliness, whereas evidence extraction remains only moderately accurate even in the oracle setting (oracle Evidence F1 between 0.47 and 0.55 across models).
    \item \textbf{Context-based retrieval outperforms full-text for hypotheses extraction.} Increasing retrieved context from $k{=}5 \rightarrow 10 \rightarrow 20$ is generally beneficial or neutral for hypothesis extraction, where as full-text prompting often underperforms context-based retrieval.
    \item \textbf{Retrieval improvements yield gains but leave substantial headroom.} Reranking and fine-tuned retrieval improve performance in both stages, but evidence extraction remains far from oracle even when the gold paragraph is provided. This shows that improving the retriever alone is not enough, the extractor itself is a significant bottleneck. 
\end{itemize}

Together, these experiments provide a controlled comparison of context selection for hypothesiscand evidence extraction in a within-document setting, including oracle contexts that separate paragraph-selection effects from extractor limitations.\footnote{\label{fn:repo}Code, prompts, and experiment configurations are available at \url{https://github.com/SaiDileepKoneru/ScientificClaimTraceExtraction.git}}

\section{Related Work}

\subsection{Statement extraction from scientific text}
Prior work on extracting scientific statements (e.g., claims and hypotheses) from scholarly text has often formulated the problem as sentence/span detection or sequence tagging. Early efforts concentrated on claim sentence identification from scientific abstracts by using discourse structure and transfer learning to distinguish claim-like sentences from surrounding context \cite{achakulvisut2019claim,jansen2016extracting}. More recent work strengthens claim sentence classifiers with objectives such as supervised contrastive learning to separate claim sentences from topically similar non-claims \cite{wei2023claimdistiller}. Complementary lines of research move beyond sentence-level detection toward structured representations of scientific statements, extracting fine-grained claim graphs that encode entities, relations, and modifiers for downstream scientific knowledge graph construction \cite{magnusson2021extracting}.

Rhetorical zoning and sequential sentence classification provide mechanisms for distinguishing statement types (e.g., background, objective, methods, results, conclusion), helping separate hypothesis/claim-like content from other discourse roles \cite{teufel2002summarizing,dernoncourt2017pubmed}. Citation-focused resources further capture how scientific assertions are situated and supported, for example through citation intent classification and related citation/discourse modeling \cite{cohan2019structural}.

A recurring theme in this literature is that moving from abstracts to full papers introduces both scale and ambiguity, as targets may be sparse, paraphrased across sections, and interleaved with rhetorically adjacent but non-target statements \cite{white2011hypothesis}. Document-level scientific information extraction benchmarks underscore the difficulty of extracting structured information when evidence spans multiple sentences or must be aggregated across a document \cite{jain2020scirex,luan2018multi}. Because scientific claims and hypotheses are embedded in structured discourse, extraction often benefits from discourse cues and argumentation structure \cite{lauscher2018arguminsci,li2021scientific,fergadis2021argumentation}. Finally, prior work also emphasizes definitional ambiguity and annotation challenges that vary across tasks and communities \cite{wu2025research,vasu2024scihyp}. 

Our work moves beyond sentence-level claim and hypothesis identification to extract structured links between findings, hypotheses, and supporting evidence in full papers. We address full-text ambiguity by systematically varying how context is constructed, separating paragraph-selection errors from extractor limitations.

\subsection{Evidence retrieval for claim verification}

A related line of work studies scientific claim verification, where systems must retrieve relevant documents and identify evidentiary sentences supporting a veracity decision. Research on general-domain fact verification and rationale evaluation uses the retrieve-then-reason paradigm with explicit evidence selection \cite{thorne2018fever,deyoung2020eraser}. In the scientific domain, this involves retrieving evidence-containing abstracts and selecting rationale sentences that support or refute a claim \cite{wadden2020fact}. Subsequent work emphasizes the importance of retrieval and context scope by decomposing the task into abstract retrieval, sentence selection, and label prediction \cite{pradeep2020scientific}, and incorporating full-document context and weak supervision to improve robustness and domain transfer \cite{wadden2021multivers}. We adopt a similar retrieve-then-extract paradigm adapted to extract structured claim traces rather than predict veracity labels. Additionally, while verification pipelines typically retrieve evidence across multiple documents, our setting targets extraction of hypothesis-linked, statistical evidence within a single document to reconstruct a paper's empirical argument.

\subsection{Impact of retrieval configuration}

Recent benchmarks for scientific long-context understanding show that providing models with long inputs alone does not guarantee reliable scientific comprehension and reasoning across tasks \cite{cui2025curie}. This motivates careful retrieval configuration because irrelevant yet semantically related passages can actively mislead downstream LLM behavior and degrade correctness in retrieval-augmented settings \cite{cuconasu2024power, amiraz2025distracting, yoran2023making}. Controlled studies demonstrate that LLMs are distractable. Adding irrelevant context can skew responses even when the prompt otherwise contains useful information, and this effect is strongest when the irrelevant context is highly similar to the target \cite{wu2024easily, shi2023large}, a scenario inherently common in within-document retrieval. Beyond relevance, the length and location of context provided also impacts performance with prior work on long-context evaluations documenting {lost-in-the-middle} behavior where models underuse relevant information buried mid-context \cite{liu2023lost}. Furthermore, studies suggest that increasing the amount of retrieved context offers diminishing returns, improving performance initially but potentially hurting it for long inputs \cite{jin2024long}. Interplay between retrieval recall and precision is also complex; subsequent filtering or selection steps may only be helpful if the initial retrieval achieves high recall \cite{li2024does}. 

Together, these findings motivate two-stage retrieval designs that prioritize high recall in the first stage of retrieval and then improve precision via reranking retrieved paragraphs into a candidate set \cite{chen2025scirerankbench,gao2023retrieval}. {We follow this approach. We first retrieve a high-recall candidate set and then use reranking to improve precision, enabling controlled comparisons of context quantity and quality in the presence of highly similar within-document distractors.}

\section{Task Description}
Our goal is to extract a structured trace of an empirical argument from the full text of a given scientific paper. A key assumption of ours is that claims are empirical in nature, so we focus exclusively on extracting statistical evidence that supports a stated hypothesis and do not consider other forms of scholarly reasoning, such as theoretical arguments.

Formally, let a scientific article $D$ be modeled as a sequence of paragraphs
$(p_1, p_2, \ldots, p_n)$, where each paragraph $p_i$ is a sequence of sentences
$p_i = (s_{i,1}, s_{i,2}, \ldots, s_{i,m_i})$. During preprocessing, we represent the article body as plain-text paragraphs only, excluding non-prose content such as tables, figures (and captions), and equations, so each $p_i$ corresponds to a contiguous block of running text from the main body. Let $F_A$ denote an \emph{abstract finding} sentence from the Abstract of $D$. Our objective is to extract two text spans from the body of $D$:

\begin{itemize}[leftmargin=*]
    \item A \emph{core hypothesis} $H$ is an interpretive or hypothesis-level statement related to the abstract finding. $H$ may be a single sentence or span multiple consecutive sentences.
    \item \emph{Statistical evidence} $E(H)$ is a text span containing statistical results (e.g., p-values, confidence intervals, regression coefficients) that substantiate the hypothesis $H$.
\end{itemize}
This task presents several unique challenges that differentiate it from standard information extraction. Notably, the link from a finding reported in the Abstract to statistical evidence is typically indirect. 
Statistical evidence 
supports a hypothesis, which in turn supports the finding. 

\section{Methodology}
\begin{figure*}
    \centering
    \includegraphics[width=0.8\linewidth]{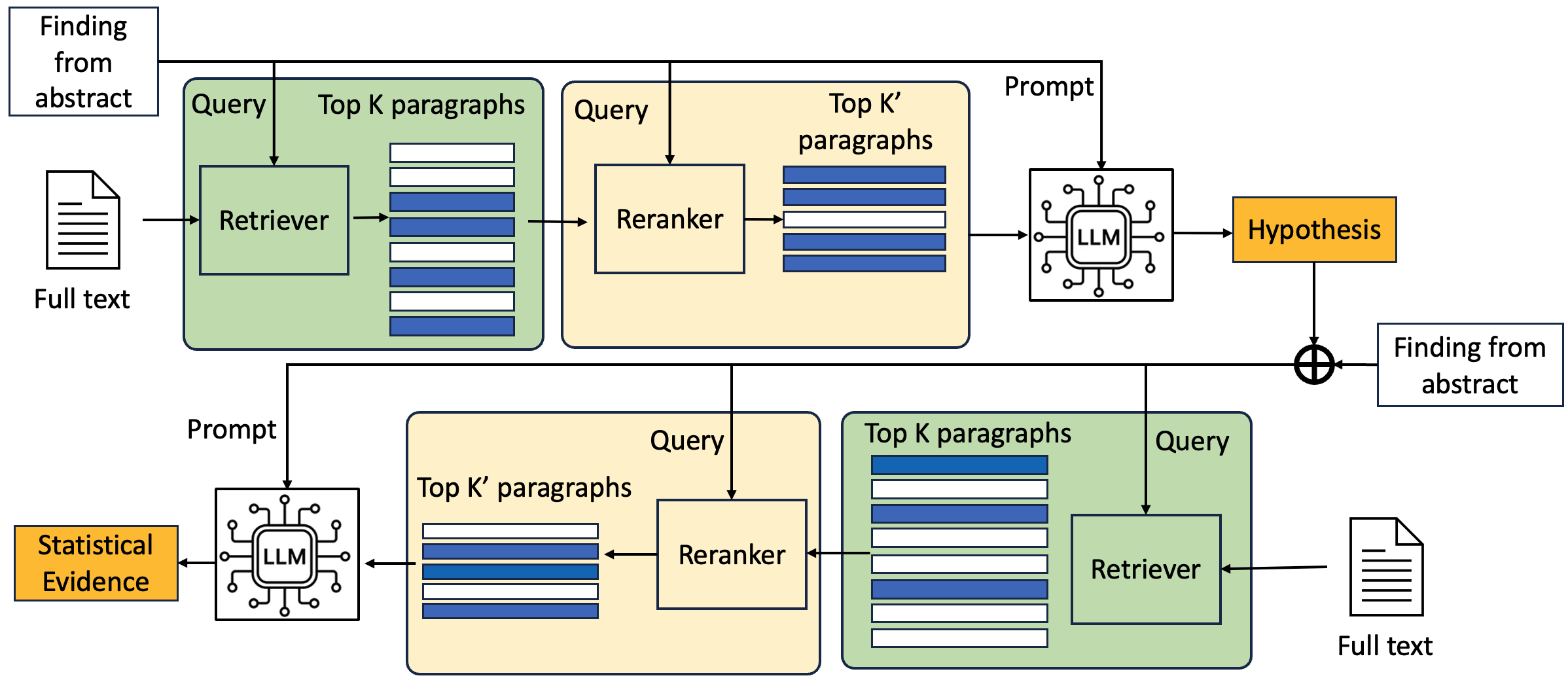}
    \caption{Schematic of the sequential retrieve and extract pipeline. Stage 1 (top) uses the abstract finding to extract the Hypothesis, which then serves as part of a composite query for extracting the Statistical Evidence in Stage 2 (bottom).}
    \label{fig:two_stage_schematic}
\end{figure*}
\subsection{Extraction pipeline}\label{subsec:pipeline}
We frame the task as a sequential, two-step process that first extracts a hypothesis and then extracts the statistical evidence supporting that hypothesis (see Figure~\ref{fig:two_stage_schematic}).

\textbf{Stage 1: Hypothesis extraction.} We use the abstract finding as a query to identify paragraphs most likely to contain the core hypothesis in the full text. We first convert the article body into paragraphs and then construct a stage-specific context set $C_1$ using one of the context configurations in Section~\ref{sec:context_configs}. Specifically, retrieval-based configurations first retrieve a candidate pool of size $M$ (e.g., $M{=}k$ for standard RAG, or $M{=}30$ when using a reranker), and optionally apply an LLM reranker (RankGPT~\cite{sun2023chatgpt}) to select a final top-$k$ context set. For reranked configurations, we use $M{=}30$ and retain $k{=}5$ paragraphs, matching the fixed context size in Table~\ref{table:results}. Full-text configurations skip retrieval and set $C_1$ to the entire article body (excluding abstract and bibliography). Given the resulting context $C_1$ and the finding $F_A$, an LLM extractor generates the hypothesis $\hat{H}$.

\textbf{Stage 2: Evidence extraction.} We form a composite query by concatenating the abstract finding with the extracted hypothesis $q_2 = [F_A;\hat{H}]$, and construct an evidence-stage context set $C_2$ using the same retrieval and reranking procedure (or full-text/{oracle context}, depending on configuration). The extractor is then prompted with $F_A$, $\hat{H}$, and $C_2$ to extract the statistical evidence $\hat{E}$. For all extraction steps, we use prompts adapted directly from the human annotator guidelines in ~\cite{alipourfard2021systematizing}. Full prompt templates (hypothesis extraction, evidence extraction, and evaluator parsing) and experiment configuration files are available in the accompanying repository.\textsuperscript{\ref{fn:repo}}%

\subsection{Context Configurations}\label{sec:context_configs}

We compare the following context construction configurations, each instantiated within the sequential two-stage pipeline: 

\textbf{Zero-Shot (Full text, two-stage):} This baseline uses the same two-stage extraction pipeline but replaces retrieved contexts with the full article body. In Stage 1, we provide the full text (excluding bibliography) and $F_A$ and prompt the LLM to extract $\hat{H}$. In Stage 2, we provide the full text again together with $F_A$ and $\hat{H}$ and prompt the LLM to extract $\hat{E}$. This isolates the effect of context selection while keeping the sequential formulation unchanged.

\textbf{CoT (Full text, two-stage):} This baseline is identical to the full-text two-stage baseline, but uses chain-of-thought prompting~\cite{kojima2022large} within each stage (i.e., the model produces intermediate reasoning before outputting $\hat{H}$ in Stage 1 and $\hat{E}$ in Stage 2). This tests whether improved prompting alone can offset the lack of context selection.

\textbf{Standard RAG:} In each stage, we use a SciBERT retriever to select the top-$k$ paragraphs from the article body and pass them to the LLM extractor. We evaluate $k \in \{5,10,20\}$.

\textbf{Standard RAG + reranker:} In each stage, we retrieve a candidate pool of $M{=}30$ paragraphs using SciBERT, then apply RankGPT to rerank and retain the top-$k$ paragraphs. We use $k{=}5$.

\textbf{Fine-tuned retriever + reranker:} This process is driven by a task-aware paragraph retriever that we fine-tuned to distinguish rhetorical function of paragraphs. The retriever is built upon a bi-encoder architecture, using SciBERT \cite{beltagy2019scibert} as a shared bi-Encoder to generate dense representations of the query and each candidate paragraph. To better distinguish the rhetorical functions between two paragraphs, the embeddings output by SciBERT are passed through two separate linear projection heads. We treat hypothesis retrieval (Stage 1) and evidence retrieval (Stage 2) as two related tasks implemented via separate projection heads over the shared SciBERT encoder. The model is fine-tuned using a multi-task triplet loss, minimizing the distance to positive paragraphs while maximizing the distance to negative ones. 

To optimize the multi-task triplet loss, we used a consistent hard negative identification strategy for both the hypothesis and evidence tasks. For each training example, the ground-truth paragraph was used as a positive point. To identify a hard negative point, we search the rest of the document for the paragraph that is most semantically similar to the positive paragraph, as measured by the cosine similarity of their SciBERT embeddings {(Appendix Figure~\ref{fig:hard_negative_example} shows an example of a selected hard negative)}. To ensure we did not select unannotated paraphrases of the hypothesis, we filtered out candidates where any constituent sentence exceeded a similarity threshold of 0.89 with the human annotated hypothesis. This threshold matches the calibrated semantic-match criterion used in hypothesis evaluation (Section~\ref{sec:evals}).

This produces challenging training examples in which the model must distinguish between highly similar paragraphs that differ in rhetorical role with respect to the query. This process was identical for both tasks, differing only in the anchor query with the abstract finding for the hypothesis task, while the composite query formed by concatenating the finding from abstract with ground truth hypothesis was used for the evidence task. During inference, we use the fine-tuned retriever with the same reranking setup as the RAG + Reranker baseline. We retrieve 30 candidates, rerank with RankGPT, and provide the top 5 paragraphs to the extractor.

\textbf{Oracle setting:} To establish an upper bound under perfect paragraph selection, we provide the extractor with gold paragraphs. In Stage 1, we set $C_1$ to the paragraph containing the human-annotated hypothesis. In Stage 2, we provide the ground-truth hypothesis (to avoid error propagation) and set $C_2$ to the paragraph containing the human-annotated statistical evidence. This isolates extractor limitations from retrieval failures.

\section{Experiments}

We conduct our experiments on the dataset introduced in \citet{alipourfard2021systematizing}, comprised of 2,257 full-text scientific articles from social and behavioral science (SBS) disciplines. Each document includes a single domain expert-annotated claim trace consisting of a high-level conceptual finding from the abstract, an interpretive or hypothesis-level statement from the body text (hypothesis), and related statistical result (evidence) \cite{fraser2023predicting}. On average, each document contains 18,268 tokens and 367 sentences after pre-processing, which involved removing the abstract and bibliography sections.

\subsection{Evaluation Metrics} \label{sec:evals}
We evaluate hypothesis and evidence extraction with metrics tailored to the properties of each output. Throughout, we report {micro-averaged} Precision/Recall/F1 computed over extracted {units} (sentences for hypotheses; structured components for evidence), aggregated over all matched units in the evaluation set.

\emph{Hypothesis Extraction (semantic match).}
Hypotheses are frequently paraphrased across a paper, so exact string overlap is overly strict. We therefore evaluate hypothesis extraction using a calibrated semantic matching criterion. We first construct a labeled calibration set by sampling model-extracted hypothesis from a held-out split and rating each extraction against the ground-truth hypothesis using a four-level rubric: \emph{exact match} (3), \emph{mostly similar} (2), \emph{partially similar} (1), and \emph{different} (0). Full annotation guidelines and examples are available in the accompanying code repository.\textsuperscript{\ref{fn:repo}} We then binarize these labels by treating scores 2-3 as Correct and 0-1 as Incorrect.

We embed predicted and gold hypothesis text using Gemini embeddings with the \texttt{SEMANTIC\_SIMILARITY} task type, which is optimized for semantic textual similarity (e.g., recommendation and duplicate-detection use cases), and compute cosine similarity. We select a similarity threshold $\tau$ on the labeled calibration set via a grid search over $\tau \in [0.1,1.0]$ with step size 0.01, choosing the value that maximizes F1 (Appendix Figure~\ref{fig:calibration_plot}). {We use the resulting threshold, 
$\tau=0.89$, as the semantic-match cutoff in all hypothesis evaluations.} At evaluation time, we split the predicted hypothesis span into sentences and count a predicted sentence as a match if its cosine similarity to the gold hypothesis (or any gold sentence, when the gold span contains multiple sentences) is at least $\tau$. Precision, recall, and F1 are then computed over matched vs.\ unmatched predicted sentences and gold sentences. This procedure rewards semantically equivalent paraphrases while penalizing additional non-hypothesis content included in the extracted span. {Appendix Figure~\ref{fig:qual_hyp_example} shows a representative extraction failure case.}

\emph{Evidence Extraction (component-level scoring).}
Statistical evidence is typically hybrid numeric-textual content, requiring evaluation that jointly measures semantic alignment and field-level correctness (variables, tests, coefficients, intervals). Exact string overlap is sensitive to rephrasing, and embedding similarity may not reflect numerical fidelity. Accordingly, we use a component-level scoring procedure consistent with structured evaluation paradigms in recent scientific benchmarks \cite{majumder2024discoverybench}.

\textbf{Component schema.} We manually inspected 50 randomly sampled gold evidence spans to identify the most consistently expressed fields in prose statistical reporting, and define a fixed set of five component types aligned with our evaluator prompts: (i) {variables} (key predictors/outcomes and constructs); (ii) {effect relationships} (e.g., directionality and significance statements); (iii) {confidence intervals}; (iv) {statistical test/model family}; and (v) {sample size / degrees of freedom} (when stated).

\textbf{Evaluator-assisted component parsing.}
Given a predicted evidence span and its corresponding gold span, we use a fixed evaluator LLM (GPT-4) with fixed, component-specific prompts to parse both texts into structured JSON fields for each component type.\textsuperscript{\ref{fn:repo}} For set-like components (variables, relationships, confidence intervals), the evaluator returns the number of items in the gold statement ($|A|$), the number of items in the extracted statement ($|B|$), and a fuzzy-matched intersection count ($|A\cap B|$). For binary components (test/model family; sample size/df), the evaluator returns a match indicator and the extracted strings from each text. For binary components, we treat {both-absent} as {not applicable}: when the evaluator returns empty strings for both the gold and predicted fields (e.g., $\texttt{gold\_test}=\texttt{predicted\_test}=\texttt{""}$), we exclude that component instance from scoring rather than counting it as an error. The evaluator is used only for parsing; scoring is computed deterministically from the returned fields.

\textbf{Aggregating precision/recall/F1.}
After parsing, each component type yields a set of extracted items (e.g., a list of variables; a list of confidence intervals) or a single attribute (e.g., test/model family; sample size/df). We compute precision/recall/F1 by counting matches over these component items. Each correctly matched component item contributes one true positive; each component item produced by the system that is not supported by the gold statement contributes one false positive; and each gold component item that is not recovered contributes one false negative. For set-like components, this corresponds to counting overlap between the parsed gold and predicted item lists (with fuzzy matching for variables and confidence intervals). For binary components (test/model family; sample size/df), we score agreement between the parsed attributes as a match and disagreement as an error. When a component is absent in both the gold and predicted evidence, we exclude it from scoring. For set-like components, absence corresponds to empty sets ($|A|=|B|=0$). For binary components (test/model family and sample size/df), absence corresponds to both extracted strings being empty (e.g., $\texttt{gold\_test}=\texttt{predicted\_test}=\texttt{""}$), in which case the instance is excluded. Otherwise, we score a binary component as one item: if both sides are present and $\texttt{match}=1$, it contributes one true positive; if gold is present and predicted is empty, it contributes one false negative; if gold is empty and predicted is present, it contributes one false positive; and if both are present but $\texttt{match}=0$, it contributes one false positive and one false negative.

This component-based evaluation assigns partial credit when an extraction correctly recovers some fields (e.g., variables and direction) but omits or misstates others (e.g., confidence intervals). Since the scorer relies on an LLM-based parser for component extraction, it depends on the evaluator model and prompts; we therefore interpret it as an automated proxy for field-level (structured) fidelity of statistical evidence extraction. {Figure~\ref{fig:evidence_component_example} provides a concrete example of the parsed components and how they contribute to partial credit.}

\section{Results}
Table~\ref{table:results} reports Precision/Recall/F1 for the two-stage pipeline across four extractors and all context configurations. We evaluate Stage 1 and Stage 2 separately. 
Unless noted otherwise, we summarize trends {averaged across extractors}.

We organize the results around three inquiries: how retrieved \textbf{context quantity} affects performance under standard RAG ($k \in \{5,10,20\}$); how retrieved \textbf{context quality} affects performance at fixed context size (standard RAG $k{=}5 \rightarrow$ RAG+reranker $k{=}5 \rightarrow$ fine-tuned retriever+reranker $k{=}5$); and what \textbf{oracle contexts} reveal about retrieval versus extraction limits. Additional diagnostics, including the per-paper transition analysis and evidence extraction conditioned on ground-truth hypotheses, are reported in Appendix (Figure~\ref{fig:evidence_transitions}, Table~\ref{table:c4_results}). All configurations use the same two-stage formulation; they differ only in how $C_1$ and $C_2$ are constructed (full text, top-$k$ retrieval, retrieve-then-rerank to a fixed $k{=}5$, or oracle gold paragraphs).

Hypothesis extraction benefits reliably from targeted context selection. Averaged across extractors, full-text prompting underperforms retrieval-based contexts (mean F1 $0.39$ vs.\ $0.43/0.47/0.49$ for standard RAG with $k{=}5/10/20$), and the best non-oracle configuration (fine-tuned retriever+reranker, $k{=}5$) reaches mean hypothesis F1 $0.50$ (about +0.11 over full text). These gains are consistent with a signal-dilution effect in full papers where many paragraphs are topically related to the abstract finding but rhetorically non-target.

Evidence extraction shows a different sensitivity to context. On average, standard RAG with a small context ($k{=}5$) underperforms full-text prompting (mean Evidence F1 $0.20$ vs.\ $0.24$), but increasing $k$ improves performance (mean Evidence F1 $0.23$ at $k{=}10$ and $0.25$ at $k{=}20$). Improving retrieval quality yields clearer gains at fixed size (RAG+Reranker $k{=}5$ F1 $0.26$, Fine-tuned Retriever+Reranker $k{=}5$ F1 $0.29$). Oracle contexts reveal a moderate ceiling (oracle Evidence F1 $0.47$-$0.55$ across models). This indicates that evidence extraction is constrained not only by retrieval, but also by extractor performance.

\subsection{Effect of retrieved context quantity ($k$)}
We first vary retrieved context quantity under standard RAG by retrieving the top-$k$ paragraphs with SciBERT ($k \in \{5,10,20\}$), holding the extractor and prompting fixed. For hypothesis extraction, performance generally increases with $k$ and can saturate for some extractors (Table~\ref{table:results}), consistent with hypotheses being paraphrased and distributed across multiple locations. For evidence extraction, increasing $k$ improves performance across extractors (Table~\ref{table:results}), though performance gains are smaller and more model-dependent than for hypothesis extraction. Importantly, full-text prompting is not equivalent to large-$k$ retrieval. For hypotheses, full-text inputs underperform even the largest retrieved contexts despite having maximal recall, suggesting signal dilution from topically similar but rhetorically non-target paragraphs. For evidence, full text can remain competitive for some extractors, indicating that broader surrounding context may sometimes help models locate and interpret statistical reporting.

\subsection{Effect of retrieval quality at fixed $k{=}5$}
We next isolate retrieval {quality} by holding the final context size fixed ($k{=}5$) and comparing three retrieval variants: Standard RAG, RAG+Reranker (retrieve 30 with SciBERT, then use RankGPT to select the top 5), and Fine-tuned Retriever+Reranker (replace SciBERT with the fine-tuned retriever while keeping the same reranking procedure). Across models, improving retrieval quality yields consistent gains for both subtasks (Table~\ref{table:results}), but with different implications.

For hypothesis extraction, reranking and fine-tuning primarily help by selecting rhetorically appropriate hypothesis statements from many topically similar within-document candidates: average F1 increases from $0.43$ (Standard RAG, $k{=}5$) to $0.47$ (RAG+Reranker) and to $0.50$ (Fine-tuned Retriever+Reranker). For evidence extraction, retrieval quality improvements also help, but the gains are consistent with improved retention of evidence-relevant paragraphs at a fixed context size. Notably, the fine-tuned retriever can also introduce trade-offs by changing which paragraphs are retained at fixed context size; we analyze these per-paper retention and loss patterns in Appendix (Figure~\ref{fig:evidence_transitions}).

{
\begin{figure}[t]
\centering
\setlength{\fboxsep}{6pt}
\setlength{\fboxrule}{0.6pt}
\fbox{
\begin{minipage}{0.97\linewidth}
\footnotesize
\textbf{Example: Component-level comparison for evidence extraction} \\

\textbf{DOI.} \href{https://doi.org/10.1016/j.jesp.2017.11.013}{10.1016/j.jesp.2017.11.013}\\

\textbf{Gold evidence (annotated).}
\emph{Out-group members were described less frequently in terms of their basic mental states (estimated marginal mean = 9.70\%) than were in-group members (estimated marginal mean = 27.60\%), $F(1,68)=14.948$, $p=.000$.} \\

\textbf{Extracted evidence.}
\emph{Participants writing about presumed out-group members referenced their complex mental states less frequently (estimated marginal mean = 0.5\%) than those writing about presumed in-group members (estimated marginal mean = 3.6\%; $F(1,68)=8.457$, $p=.005$).} \\

\textbf{Parsed component overlap (evaluator parses; scoring is deterministic).}\\[-2pt]
\begin{center}
\resizebox{\linewidth}{!}{
\begin{tabular}{lcc}
\hline
\textbf{Component} & \textbf{Overlap / Match} & \textbf{Notes} \\
\hline
Variables (set) & $|A\cap B|=2$ (of $|A|=3$, $|B|=7$) & partial overlap \\
Relationships (set) & $|A\cap B|=2$ (of $|A|=2$, $|B|=5$) & extra predicted relations \\
Confidence intervals (set) & NA (absent in both) & not scored \\
Test / model family (binary) & $1$ & F-test family \\
Sample size / df (binary) & $1$ & $df=68$ \\
\hline
\end{tabular}}
\end{center}

\end{minipage}
}
\caption{{Illustration of component-level evaluation of evidence extraction. The metric assigns partial credit when the extracted span captures the relationship but corresponds to a different reported effect with different numeric details.}}
\label{fig:evidence_component_example}
\end{figure}
}

\subsection{Oracle contexts determine extraction limits}
Finally, we evaluate an oracle context setting where the extractor is given the gold paragraph containing the human-annotated hypothesis or evidence. This removes retrieval failures and provides an upper bound on extraction performance under perfect paragraph selection.

For hypothesis extraction, oracle contexts substantially improve performance across all extractors (mean Hypothesis F1 $0.73$), leaving a large gap relative to the best non-oracle configuration (Fine-tuned Retriever+Reranker, mean F1 $0.50$). This indicates that hypothesis extraction is strongly constrained by context selection, but that extraction still has substantial headroom even when the correct paragraph is provided.

For evidence extraction, oracle contexts reveal a different bottleneck: even with perfect paragraph selection, performance remains moderate (oracle Evidence F1 $0.47$-$0.55$ across models; Table~\ref{table:results}). This upper bound likely reflects a mix of extractor performance for hybrid numeric-textual statements and the fact that some statistical details may be reported outside prose (e.g., in tables/figures), which we remove during preprocessing. The resulting oracle gap (e.g., Fine-tuned Retriever+Reranker $0.29$ vs.\ oracle $0.52$ on average) suggests that evidence errors are driven not only by retrieval, but also by extractor fidelity for structured statistical content (e.g., preserving variables, directionality, and numeric fields).

\emph{Diagnostic analyses.} To contextualize the aggregate scores in Table~\ref{table:results}, we provide detailed diagnostics {in the Appendix \ref{hypothesisappendix}}. First, we analyze the relevant sentence proportion (RSP) to quantify the density of hypothesis-relevant information. Across configurations, we find that higher signal density strongly correlates with improved F1 (Appendix Figure~\ref{fig:purity_bins}). Second, for evidence extraction conditioned on ground-truth hypotheses, we report a per-paper retrieval-outcome breakdown showing that improvements from retrieval changes are concentrated in cases where the gold evidence paragraph is newly retrieved, while regressions arise when reranking or retriever changes fail to retain that paragraph (Appendix Figure~\ref{fig:evidence_transitions}, Table~\ref{table:c4_results}).

\begin{table*}[h]
\fontsize{8}{11}
\selectfont{
\centering
\begin{tabular}{|l|l|lll|lll|}
\hline
\textbf{LLM} & \textbf{Method} & \multicolumn{3}{c|}{\textbf{Hypothesis }} & \multicolumn{3}{c|}{\textbf{Evidence}}\\
& & P & R & F1 & P & R & F1 \\
\hline
\multirow{8}{*}{llama 3.2} & Zero-Shot (Full text) & 0.34 & 0.40 & 0.36 & 0.18 & 0.20 & 0.19\\
&CoT (Full text) & 0.21 & 0.34 & 0.26 & 0.19 & 0.22 & 0.20 \\
\cline{2-8}
&RAG (k=5) & 0.45 & 0.36 & 0.40 & 0.17 & 0.22 & 0.19\\
&RAG (k=10) & 0.50 & 0.40 & 0.44 & 0.18 & 0.22 & 0.20\\
&RAG (k=20) & 0.50 & 0.41 & 0.45 & 0.20 & 0.21 & 0.21\\
&RAG + Reranker (k=5) & 0.45 & 0.41 & 0.43 & 0.19 & 0.24 & 0.21\\
\cline{2-8}
&Fine-tuned SciBERT + Reranker (k=5) & \textbf{0.51} & \textbf{0.43} & \textbf{0.47} & \textbf{0.25} & \textbf{0.28} & \textbf{0.26}\\
& Oracle context & 0.67 & 0.55 & 0.61 & 0.43 & 0.52 & 0.47\\
\hline
\multirow{8}{*}{gpt-oss} & Zero-Shot (Full text) & 0.48 & 0.40 & 0.44 & 0.23 & 0.27 & 0.24\\
&CoT (Full text) & 0.41 & 0.50 & 0.45 & 0.25 & 0.28 & 0.26\\
\cline{2-8}
& RAG (k=5) & 0.52 & 0.43 & 0.47 & 0.15 & 0.20 & 0.17 \\
&RAG (k=10) & 0.53 & 0.46 & 0.49 & 0.19 & 0.25 & 0.21\\
&RAG (k=20)  & 0.56 & 0.50 & 0.53 & 0.22 & 0.31 & 0.25 \\
&RAG + Reranker (k=5) & 0.54 & 0.50 & 0.52 & 0.24 & 0.28 & 0.25\\ 
\cline{2-8}
&Fine-tuned SciBERT + Reranker (k=5) & \textbf{0.58} & \textbf{0.51} & \textbf{0.54} & \textbf{0.25} & \textbf{0.31} & \textbf{0.27}\\
& Oracle context & 0.91 & 0.73 & 0.81 & 0.50 & 0.57 & 0.53\\
\hline
\multirow{8}{*}{gpt-4o-mini} & Zero-Shot (Full text) & 0.33 & 0.40 & 0.36 & 0.17 & 0.22 & 0.18\\
&CoT (Full text) & 0.43 & 0.42 & 0.42 & 0.21 & 0.26 & 0.23\\
\cline{2-8}
& RAG (k=5) & 0.50 & 0.46 & 0.48 & 0.23 & 0.29 & 0.26 \\
&RAG (k=10) & 0.57 & 0.47 & 0.52 & 0.26 & 0.33 & 0.28\\
&RAG (k=20)  & 0.58 & 0.47 & 0.52 & 0.28 & 0.35 & 0.29 \\
&RAG + Reranker (k=5) & 0.56 & 0.46 & 0.50 & 0.30 & 0.39 & 0.33\\
\cline{2-8}
&Fine-tuned SciBERT + Reranker (k=5) & \textbf{0.59} & \textbf{0.49} & \textbf{0.54} & \textbf{0.33} & \textbf{0.41} & \textbf{0.35}\\

& Oracle context & 0.88 & 0.70 & 0.78 & 0.49 & 0.63 & 0.55\\
\hline
\multirow{8}{*}{gemini-2.5-flash} & Zero-Shot (Full text) & 0.35 & 0.47 & 0.40 & 0.31 & 0.48 & 0.35\\
&CoT (Full text) & 0.35 & 0.48 & 0.40 & \textbf{0.32} & \textbf{0.49} & \textbf{0.36}\\
\cline{2-8}
& RAG (k=5) & 0.40 & 0.37 & 0.38 & 0.15 & 0.20 & 0.16\\
& RAG (k=10) & 0.42 & 0.41 & 0.42 & 0.19 & 0.25 & 0.21\\
& RAG (k=20) & 0.43 & 0.45 & 0.44 & 0.22 & 0.30 & 0.25\\
& RAG + Reranker (k=5) & 0.44 & 0.43 & 0.44 & 0.24 & 0.33 & 0.27\\
\cline{2-8}
&Fine-tuned SciBERT + Reranker (k=5) & \textbf{0.46} & \textbf{0.47} & \textbf{0.46} & 0.27 & 0.39 & 0.30\\
& Oracle context & 0.74 & 0.69 & 0.71 & 0.51 & 0.59 & 0.53\\
\hline
\end{tabular}
\caption{Results for sequential two-stage extraction of hypothesis and corresponding evidence from full text of scientific articles. We compare hypothesis and statistical evidence extraction performance (Precision, Recall, F1) across four LLMs, comparing full-text baselines against retrieval-augmented strategies with varying context quantity ($k$) and retrieval quality. Oracle settings indicate performance upper bounds given perfect paragraph selection. {Best non-oracle F1 per model and subtask is bolded, and oracle results are reported as upper bounds.}}
\label{table:results}}
\end{table*}
\section{Discussion}

A defining feature of our task is that retrieval is performed {within} a single paper. Candidate paragraphs are therefore often topically related to the abstract finding, yet vary widely in rhetorical function (e.g., background framing, theoretical motivation, hypothesized relationships, results summaries, and implications). In this setting, improvements from reranking and the fine-tuned retriever are consistent with retrieval changes affecting which paragraphs appear in the final context under a fixed size. While we do not make mechanistic claims about model behavior, the aggregate patterns indicate that retrieval improvements translate more reliably into gains for hypothesis extraction than for evidence extraction.

To complement aggregate metrics, we report two descriptive diagnostics (Appendix). First, we analyze \emph{relevant sentence proportion} (RSP) as a proxy for context signal density for hypotheses: for a given hypothesis-stage context, RSP is the fraction of its sentences whose semantic similarity to the gold hypothesis exceeds the same calibrated cutoff used in our hypothesis evaluation. Higher RSP is consistently associated with higher hypothesis extraction F1 across configurations (Appendix Figure~\ref{fig:purity_bins}). Second, for evidence extraction conditioned on ground-truth hypotheses, we perform a per-paper transition analysis that buckets papers by whether a retrieval change \textsc{Gains}, \textsc{Keeps}, \textsc{Loses}, or \textsc{Misses} the gold evidence paragraph, and report the corresponding mean per-paper $\Delta$F1 (Appendix Figure~\ref{fig:evidence_transitions}, Table~\ref{table:c4_results}). These analyses describe how retrieval changes alter the retrieved contexts and where gains are concentrated (e.g., cases where the gold paragraph is newly retrieved), without attributing performance changes to specific internal model mechanisms.

Oracle contexts further separate retrieval failures from extraction limits. For hypotheses, the remaining gap between the best non-oracle setting and oracle performance indicates substantial headroom even when the correct paragraph is provided. For evidence, oracle performance remains moderate across models, which is consistent with evidence extraction requiring faithful handling of hybrid numeric-textual statements. In practice, improving this stage may require extraction methods that more strongly enforce structure (e.g., schema-constrained outputs) or hybrid approaches that combine retrieval with deterministic parsing for quantities such as coefficients, confidence intervals, and p-values.

The diagnostics and analyses above  suggest two best practices for building claim-trace extraction systems. First, for hypothesis extraction, investing in retrieval quality and retrieving moderately larger contexts are likely to significantly improve the performance. On the contrary, for evidence extraction, retrieval quality and context size matter, but gains remain limited without corresponding improvements in extractor fidelity for structured statistical content. Together, these findings emphasize that full-text extraction pipelines should treat retrieval and extraction as coupled components: retrieval is a dominant factor for hypothesis extraction, while evidence extraction additionally depends on accurately preserving and aligning statistical fields.

\section{Conclusion}
We study a sequential full-text claim-trace extraction problem, 
where an abstract-level finding is linked to (i) a corresponding hypothesis statement in the paper body and (ii) the statistical evidence that supports that hypothesis. This formulation creates a challenging {within-document} retrieval setting: candidate paragraphs are often topically related to the finding but differ in rhetorical role, and errors can propagate across stages.

Using a consistent two-stage retrieve-and-extract pipeline, we present a controlled study of how retrieval configuration affects each stage, varying retrieved context quantity (top-$k \in \{5,10,20\}$), retrieval quality (standard RAG, reranking, and a fine-tuned retriever paired with reranking), and an oracle paragraph setting to separate retrieval failures from extraction limits. Across four LLM extractors, hypothesis extraction benefits reliably from targeted context selection: retrieval-based contexts outperform full-text prompting, and improving retrieval quality yields consistent gains. Evidence extraction improves more modestly with retrieval changes, and oracle contexts remain only moderately accurate, indicating that paragraph selection alone does not eliminate evidence extraction errors for structured statistical content.

Together, these results highlight a task-dependent bottleneck in sequential full-text extraction: hypothesis extraction is strongly affected by context selection in this within-document setting, while evidence extraction additionally depends on extraction fidelity for hybrid numeric-textual statements even when the correct paragraph is available. Future work may extend this analysis beyond SBS articles and develop extraction approaches that better preserve variable binding and numerical fields in statistical evidence.

\section*{Limitations and Ethical Considerations}
Our experiments are limited to claim traces extracted from empirical research in the SBS disciplines. Writing conventions vary across disciplines, so observed trade-offs may not transfer directly to other domains. Our dataset contains a single annotated claim trace per paper and is not exhaustive, as papers often contain multiple hypotheses and supporting results.

Our preprocessing uses prose-only paragraphs and excludes tables, figures (and captions), and equations; evidence extraction results therefore reflect performance on prose-reported statistical evidence only. Accordingly, oracle Evidence F1 should be interpreted as an upper bound under perfect paragraph selection within prose, rather than a ceiling on recovering statistical details that may be reported outside running text. Furthermore, the task is sequential, so errors can propagate from Stage~1 (hypothesis extraction) to Stage~2 (evidence extraction); we partially isolate the evidence stage by evaluating evidence extraction conditioned on ground-truth hypotheses (Appendix Table~\ref{table:c4_results}). Finally, the evaluation depends on automated proxies: hypothesis scoring uses a calibrated embedding-similarity threshold, and evidence scoring uses an LLM-based component parser with deterministic scoring. We interpret these metrics as scalable approximations of semantic match (hypotheses) and field-level performance (evidence).

This work does not involve human subjects or personal data. The main risks are practical: full-text scientific articles may be subject to copyright and licensing restrictions, so we release only code, prompts, and configurations rather than redistributing full texts. In addition, hypothesis and evidence extractions can be plausible but incorrect—especially for numeric-text evidence—so outputs should be treated as assistive and audited in high-stakes use. Finally, our evidence scoring relies on an LLM-based parser and our systems depend on pretrained models, both of which may introduce systematic errors or bias and may favor dominant writing styles.

\bibliographystyle{ACM-Reference-Format}
\bibliography{main}

\appendix 
\label{sec:appendix}
\begin{figure*}[t]
    \centering
    \subfloat[LLaMA-3.2]{%
        \includegraphics[width=0.48\linewidth]{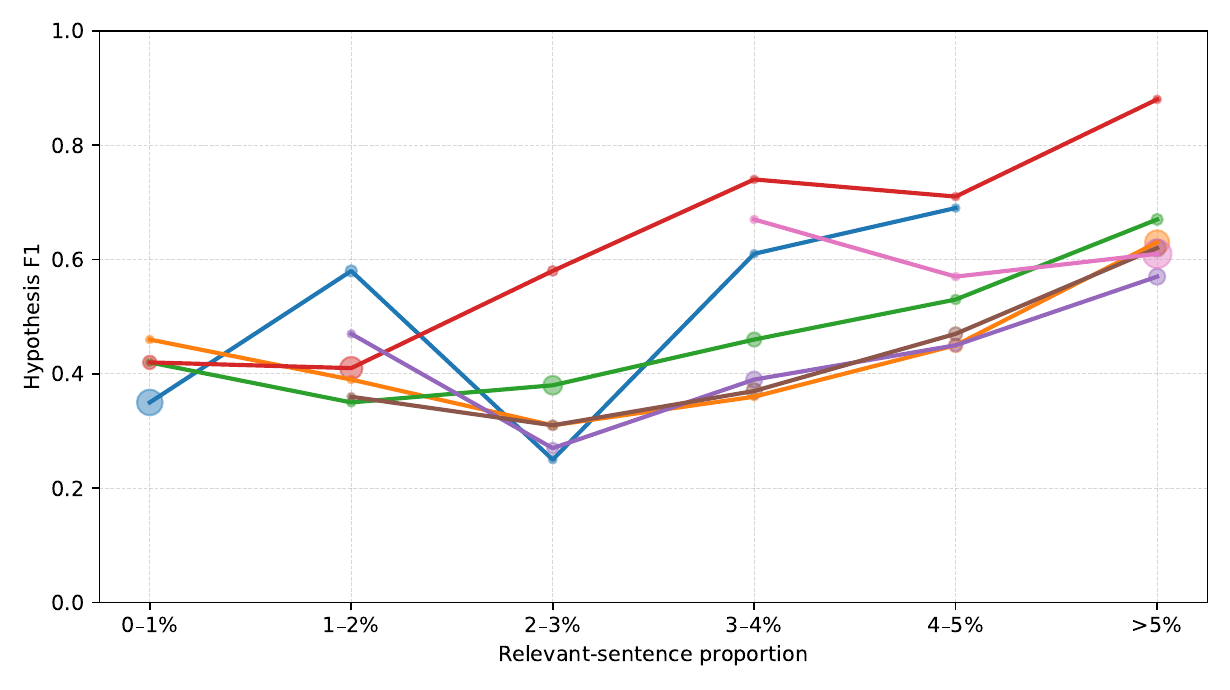}%
    }\hfill
    \subfloat[Gemini-2.5-Flash]{%
        \includegraphics[width=0.48\linewidth]{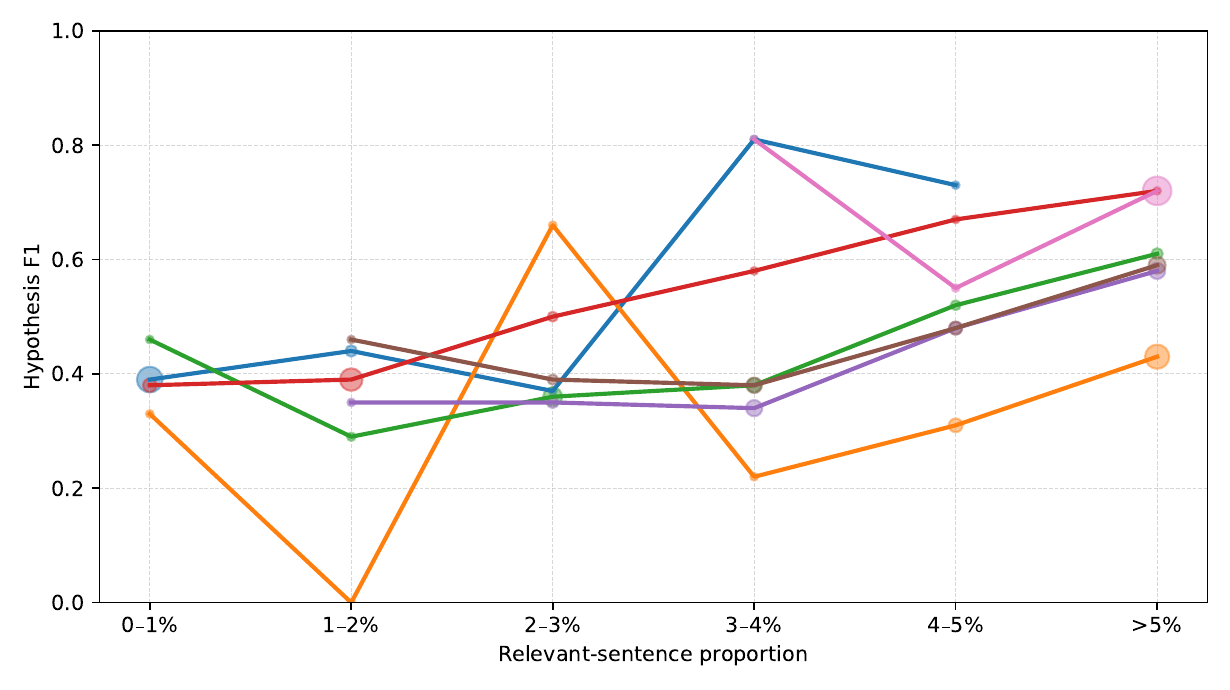}%
    }
    
    \vspace{1em} 

    \subfloat[GPT-4o-mini]{%
        \includegraphics[width=0.48\linewidth]{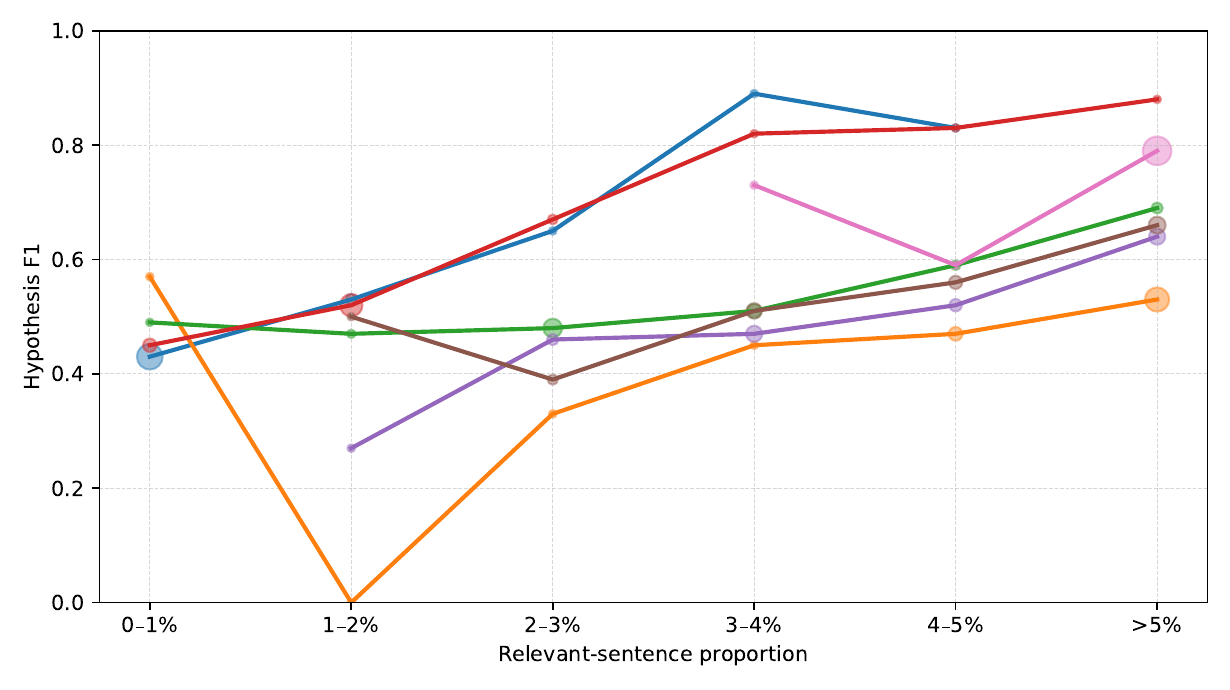}%
    }\hfill
    \subfloat[GPT-OSS]{%
        \includegraphics[width=0.48\linewidth]{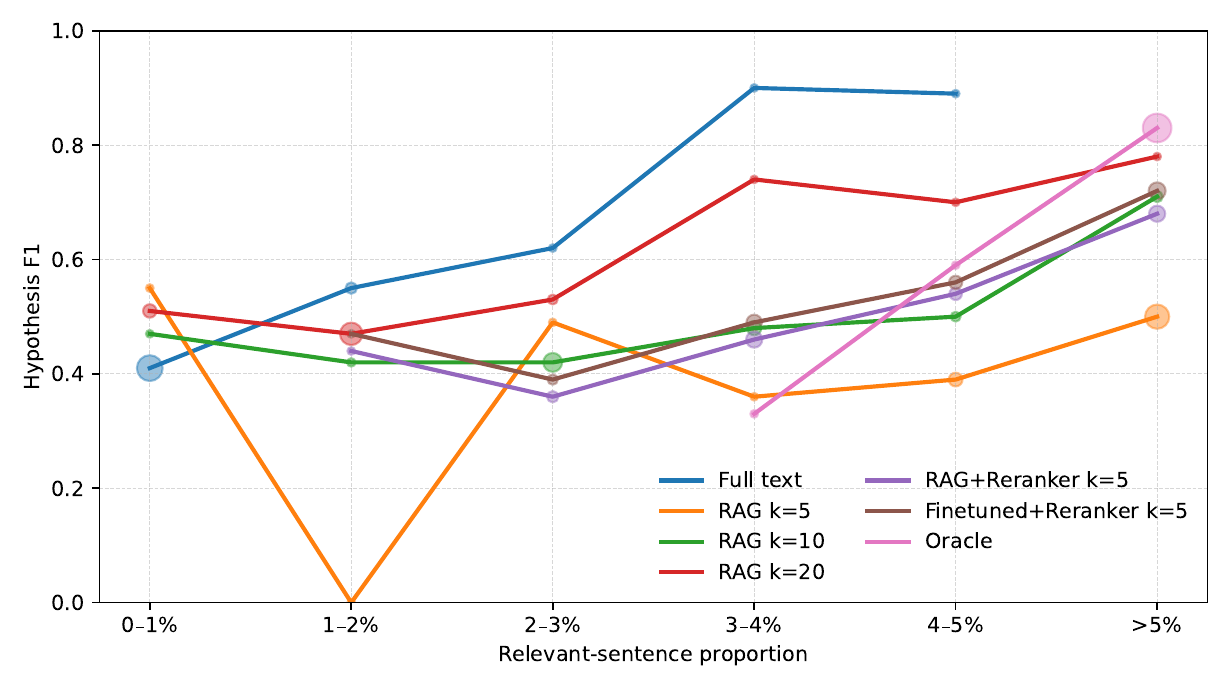}%
    }

    \caption{Hypothesis extraction F1 versus relevant-sentence proportion (RSP), binned. Points show mean F1 within each bin and point size is proportional to the number of papers in the bin.}
    \label{fig:purity_bins}
\end{figure*}

\section{Diagnostic: Context signal density (hypothesis extraction)}\label{hypothesisappendix}
For each hypothesis-stage context, we compute the \emph{relevant sentence proportion} (RSP), defined as the fraction of sentences whose cosine similarity to the gold hypothesis exceeds $\tau{=}0.89$ using Gemini embeddings and the same threshold used for hypothesis evaluation in Section~\ref{sec:evals}. We bin papers by RSP (0--1\%, 1--2\%, 2--3\%, 3--4\%, 4--5\%, and $>5$\%) and report mean hypothesis extraction F1 within each bin in Figure~\ref{fig:purity_bins}. Point size is proportional to the number of papers in the bin.

\section{Diagnostic: Per-paper transition analysis: quantity versus quality (evidence extraction)}
To isolate how retrieved context affects evidence extraction, independent of errors from the hypothesis stage, we condition on the ground-truth hypothesis and vary only the evidence-stage context construction. We compare transitions that change context quantity (standard RAG with $k\in\{5,10,20\}$ and full text) and transitions that change context quality while holding the final context size fixed (dense $k{=}5 \rightarrow$ reranked $k{=}5 \rightarrow$ fine-tuned-retriever+$k{=}5$). 

\begin{figure*}[t]
  \centering

  \subfloat[GPT-4o-mini (Quantity)\label{fig:evtrans_gpt4omini_qty}]{%
    \includegraphics[width=0.48\textwidth]{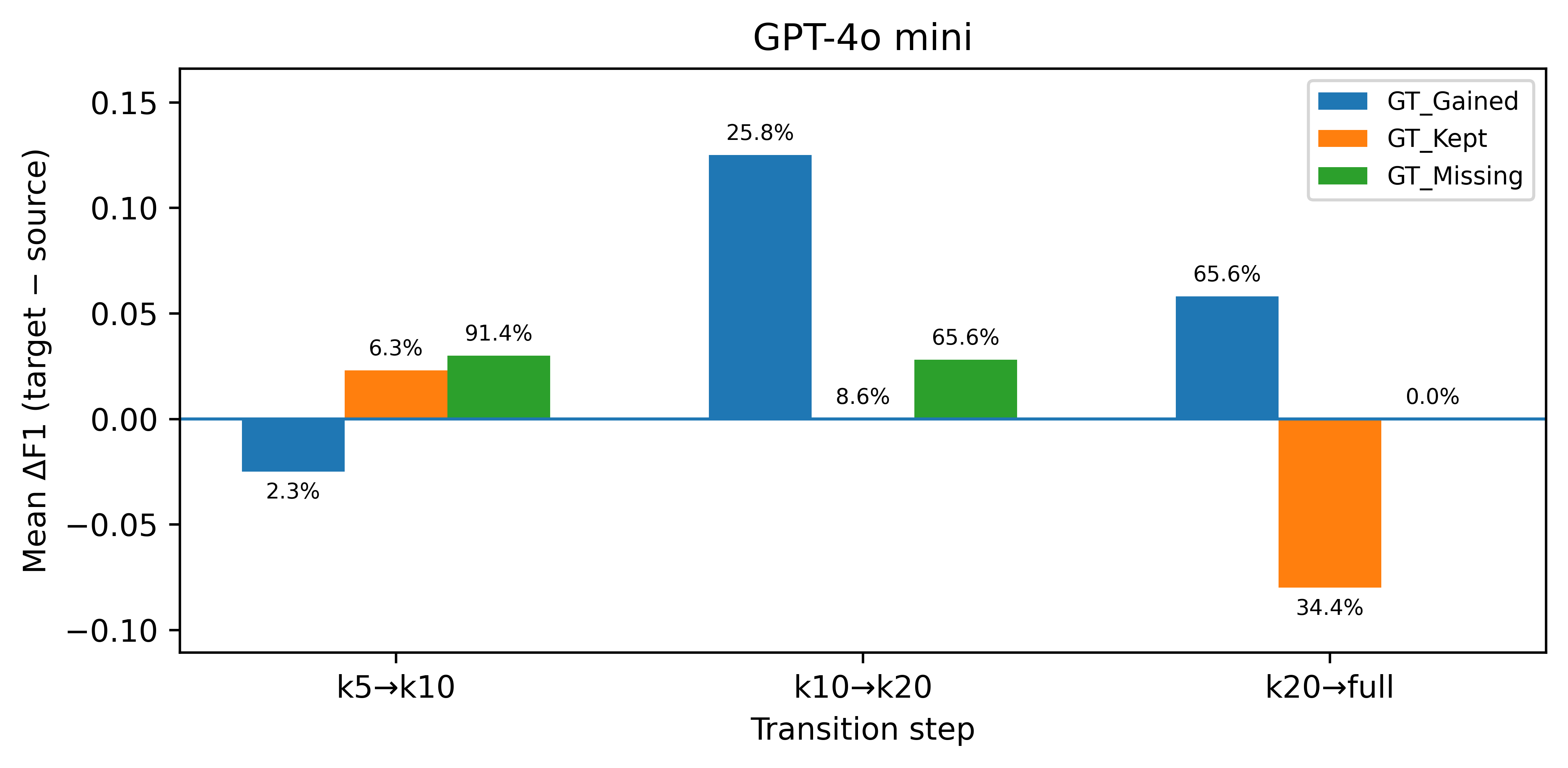}%
  }\hfill
  \subfloat[GPT-4o-mini (Quality)\label{fig:evtrans_gpt4omini_qual}]{%
    \includegraphics[width=0.48\textwidth]{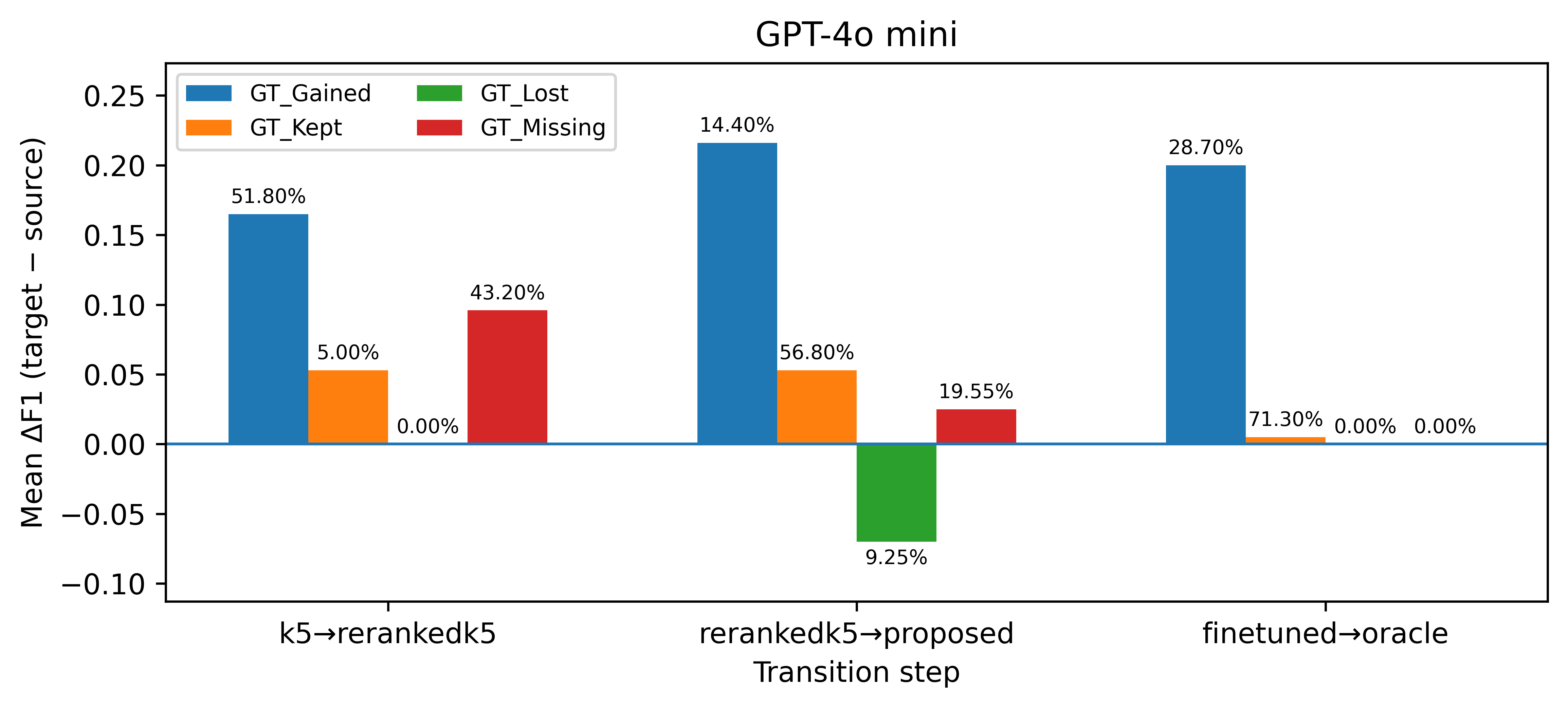}%
  }
  \vspace{0.6em}

  \subfloat[Gemini-2.5-Flash (Quantity)\label{fig:evtrans_gemini_qty}]{%
    \includegraphics[width=0.48\textwidth]{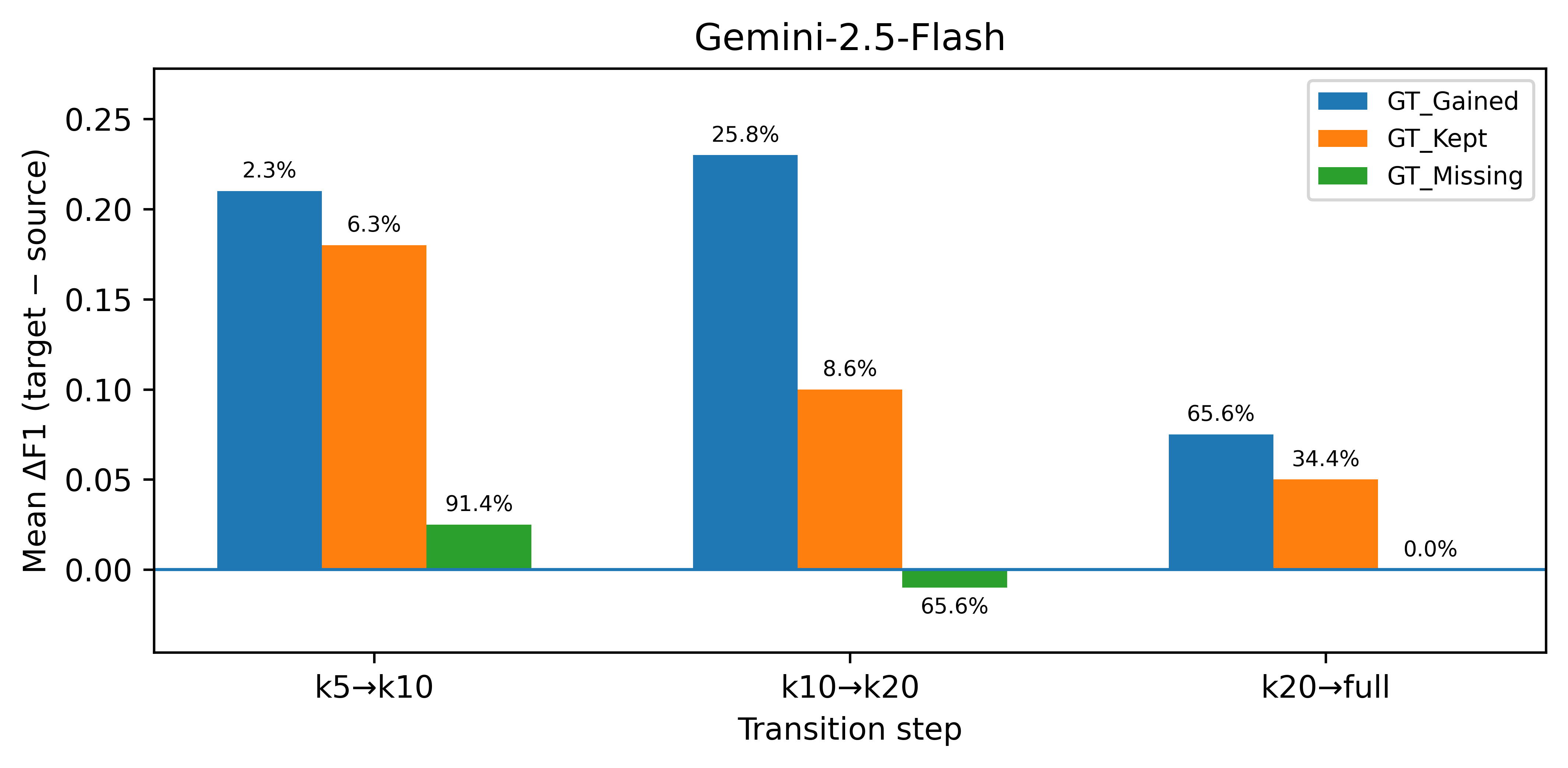}%
  }\hfill
  \subfloat[Gemini-2.5-Flash (Quality)\label{fig:evtrans_gemini_qual}]{%
    \includegraphics[width=0.48\textwidth]{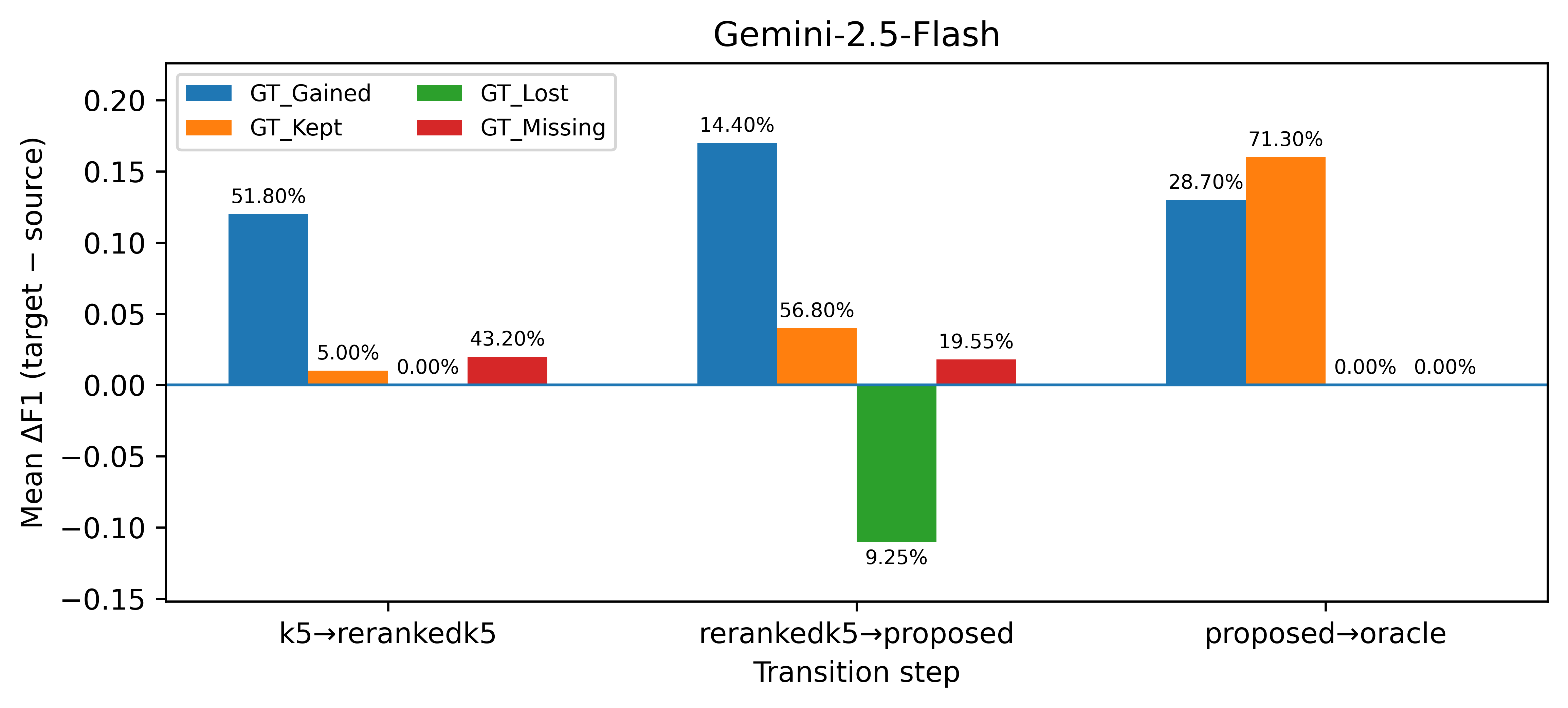}%
  }

  \vspace{0.6em}

  \subfloat[LLaMA-3.2 (Quantity)\label{fig:evtrans_llama_qty}]{%
    \includegraphics[width=0.48\textwidth]{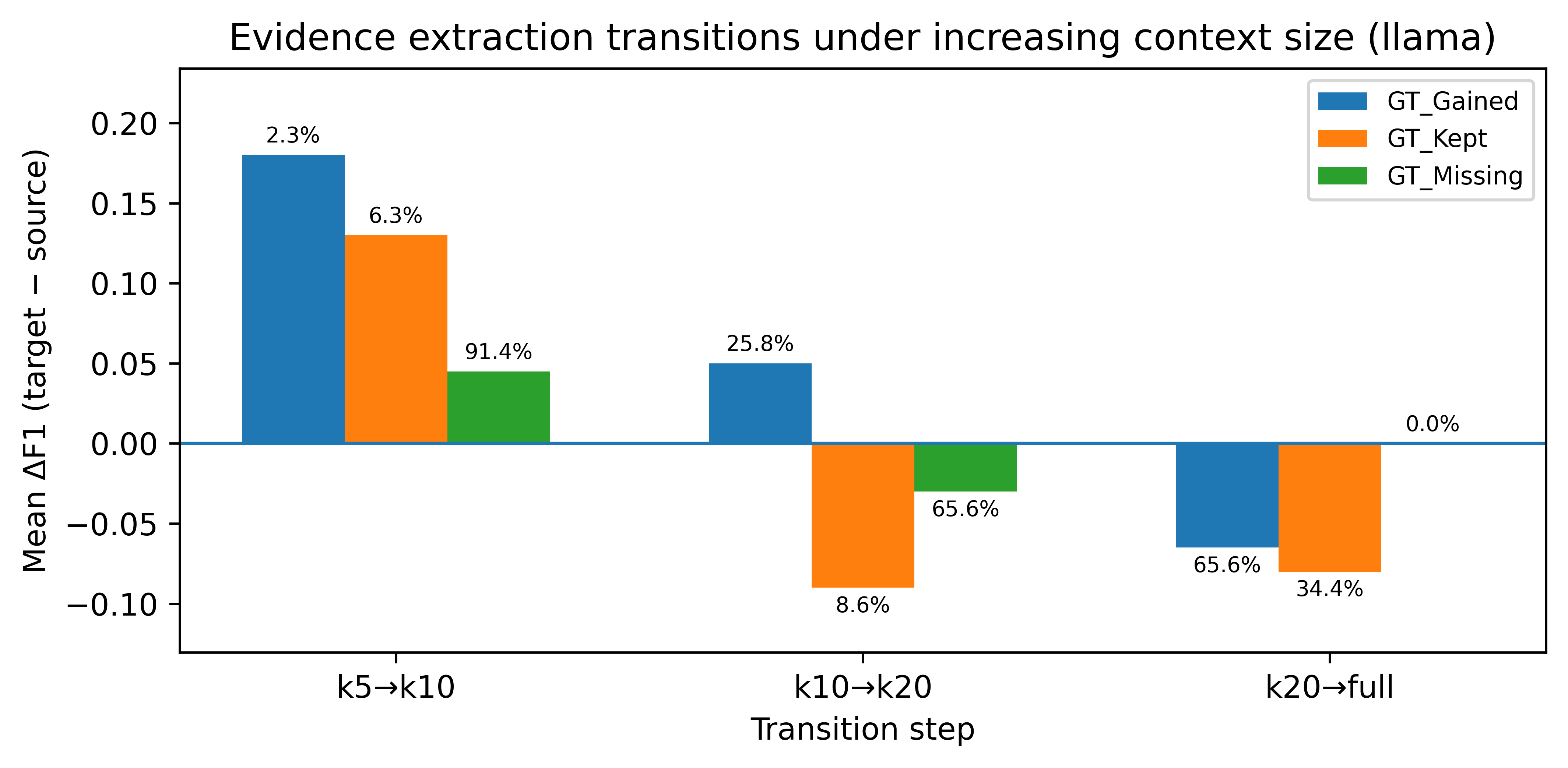}%
  }\hfill
  \subfloat[LLaMA-3.2 (Quality)\label{fig:evtrans_llama_qual}]{%
    \includegraphics[width=0.48\textwidth]{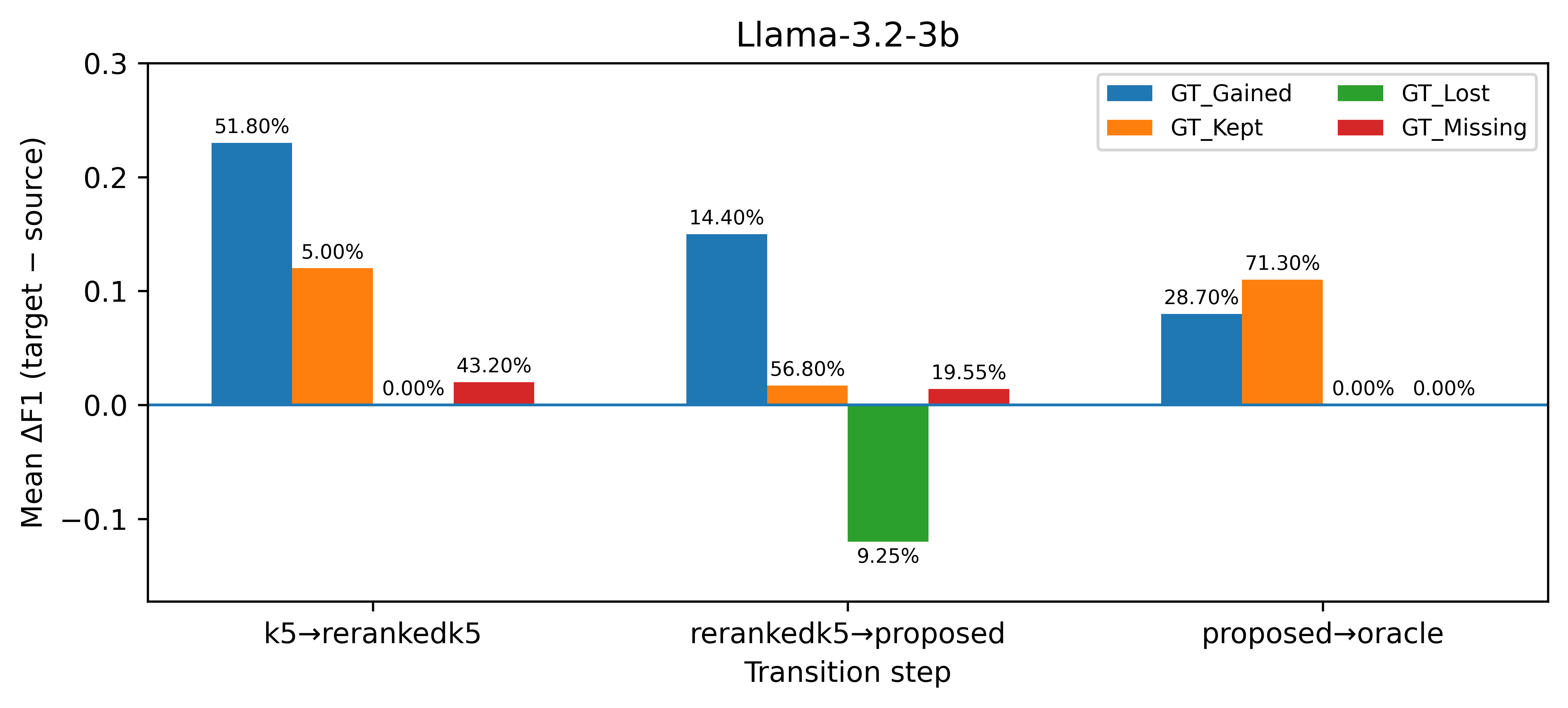}%
  }

  \vspace{0.6em}

  \subfloat[GPT-OSS (Quantity)\label{fig:evtrans_gptoss_qty}]{%
    \includegraphics[width=0.48\textwidth]{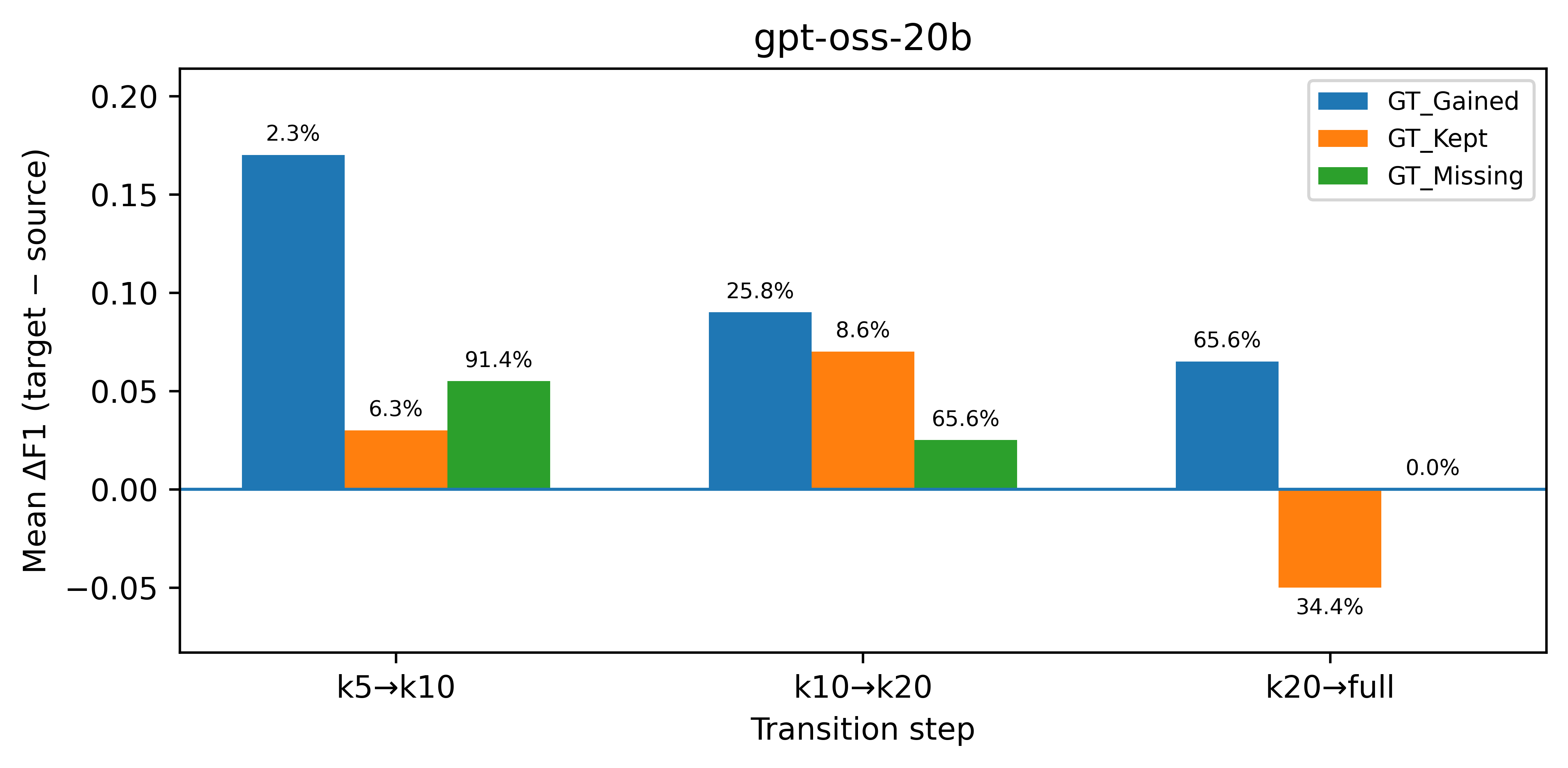}%
  }\hfill
  \subfloat[GPT-OSS (Quality)\label{fig:evtrans_gptoss_qual}]{%
    \includegraphics[width=0.48\textwidth]{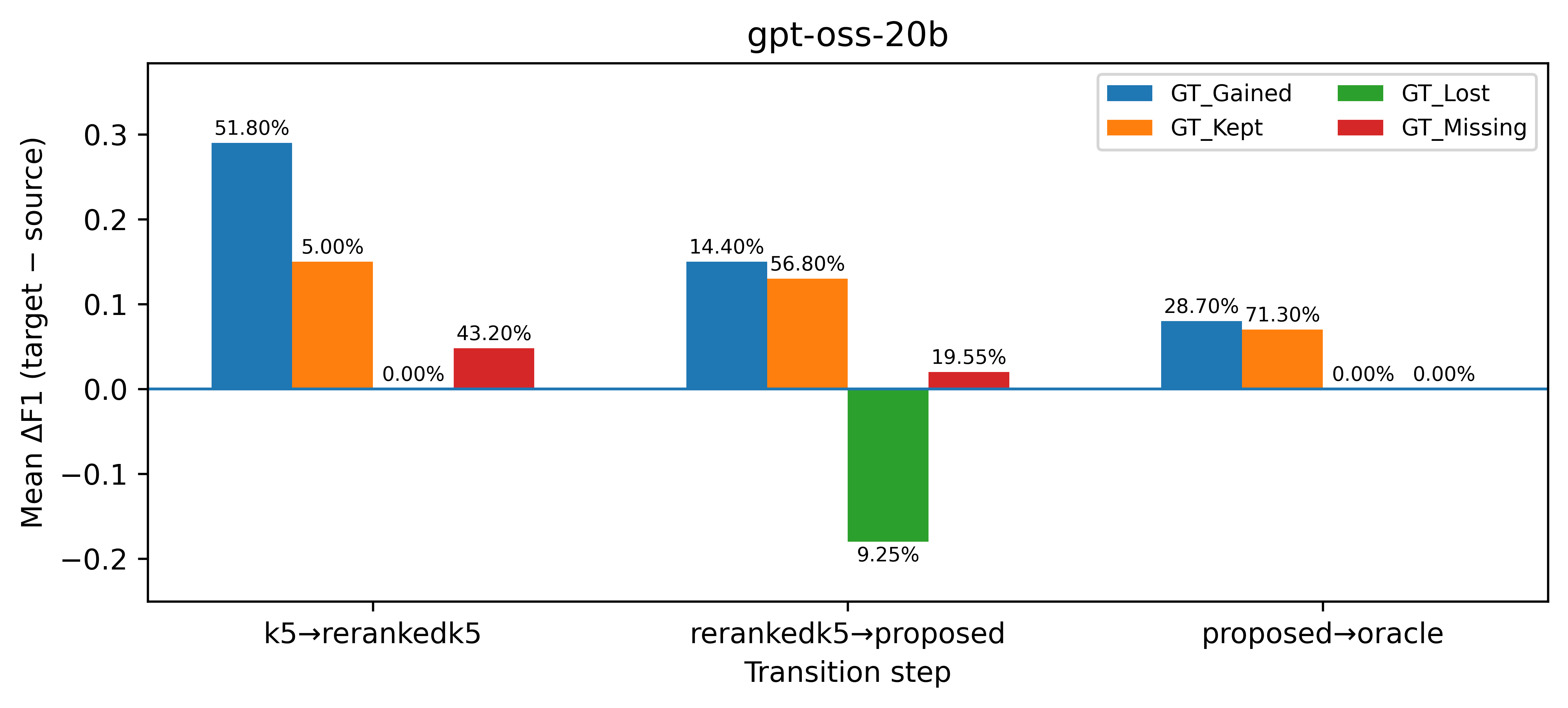}%
  }

  \caption{{Per-paper transition analysis for evidence extraction under quantity vs.\ quality changes.}
  Left column varies retrieved context quantity ($k{=}5\rightarrow 10\rightarrow 20\rightarrow$ full text). Right column varies retrieval quality at fixed context size ($k{=}5$) (RAG $\rightarrow$ RAG+Reranker $\rightarrow$ fine-tuned retriever+reranker $\rightarrow$ oracle).
  Bars report mean per-paper $\Delta$F1 within each retrieval-outcome bucket; labels above bars show the fraction of papers in that bucket. For quality transitions, we additionally report \textsc{GT\_Lost} when the gold evidence paragraph is present in the source context but absent in the target.}
  \label{fig:evidence_transitions}
\end{figure*}

For each transition (source $\rightarrow$ target), we categorize each paper by whether the gold evidence paragraph $g$ appears in the retrieved evidence-stage contexts $C_{\text{src}}$ and $C_{\text{tgt}}$. We label a paper as \textsc{GT\_Kept} if $g$ appears in both contexts ($g \in C_{\text{src}}$ and $g \in C_{\text{tgt}}$), and as \textsc{GT\_Gained} if $g$ is absent in the source but appears in the target ($g \notin C_{\text{src}}$ and $g \in C_{\text{tgt}}$). For transitions where the target context is not guaranteed to contain the source context (e.g., reranking or switching retrievers), we additionally track \textsc{GT\_Lost} cases where $g$ is present in the source but not retained in the target ($g \in C_{\text{src}}$ and $g \notin C_{\text{tgt}}$), distinguishing these from \textsc{GT\_Missing} cases where $g$ is absent from both contexts ($g \notin C_{\text{src}}$ and $g \notin C_{\text{tgt}}$). Within each bucket, we compute the mean per-paper change in Evidence F1, $\Delta\text{F1} = \text{F1}_{\text{tgt}} - \text{F1}_{\text{src}}$. We summarize the per-paper transition outcomes in Figure~\ref{fig:evidence_transitions}, and report the corresponding aggregate Precision/Recall/F1 for the same ground-truth-hypothesis setting in Table~\ref{table:c4_results}.

We find that changes in evidence extraction performance are largely explained by whether a retrieval change adds the gold evidence paragraph to the context. In quantity ablations, improvements from larger $k$ occur primarily in \textsc{GT\_Gained} cases (where expanding $k$ newly retrieves $g$). When gold paragraph is already present (\textsc{GT\_Kept}), expanding context (e.g., $k{=}20 \rightarrow$ full text) yields limited gains and can even degrade performance.

In contrast, retrieval-quality improvements (reranking and fine-tuning) yield more consistent gains because they increase the rate of \textsc{GT\_Gained} outcomes under a fixed context size. However, quality changes are not monotonic: fine-tuning can introduce a small trade-off in \textsc{GT\_Lost} cases, where $g$ was retrieved under the prior method but is not retained after the retriever/reranker change.

\section{Semantic similarity calibration}\label{subsec:calibration}
\begin{figure}[H]
    \centering
    \includegraphics[width=\linewidth]{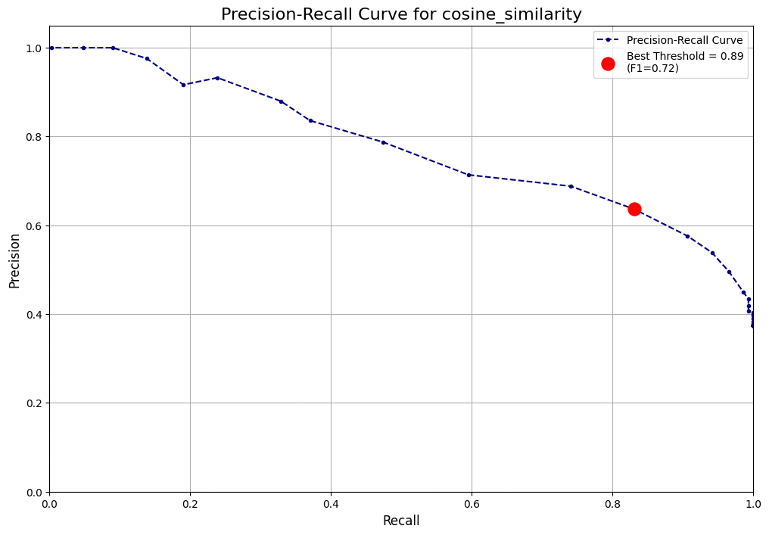}
    \caption{Calibration of the cosine similarity threshold using a precision recall curve. The plot shows the values at different threshold values for the Gemini embedding model. The selected threshold of 0.89 corresponds to the point of maximum F1 score (0.72) on manually labeled set.}
    \label{fig:calibration_plot}
\end{figure}

\begin{figure*}[t]
\centering
\setlength{\fboxsep}{6pt}
\setlength{\fboxrule}{0.6pt}
\fbox{
\begin{minipage}{0.97\linewidth}
\footnotesize
\textbf{Example: Hypothesis extraction error with high embedding similarity} \hfill
\textbf{Cosine sim:} 0.875 (87.5\%) \\

\textbf{DOI} \href{https://doi.org/10.1016/j.jesp.2017.07.008}{10.1016/j.jesp.2017.07.008}\\

\textbf{Human annotation.}
\emph{The presence of an \textbf{alternative option} will produce a competing, covert influence on participants.} \\

\textbf{Extracted.}
\emph{We hypothesize that participants will be more likely to respond opposite to the literal content of \textbf{political conspiracy statements} because competing covert influences---such as threat-protection needs and accuracy biases---drive a tension between appealing and unappealing aspects of conspiracies, resulting in opposite responses relative to baseline statements.} \\

\textbf{Diagnosis.}
High lexical/semantic overlap (``covert influence'') but the prediction shifts the core construct and mechanism from \textbf{alternative option} to \textbf{conspiracy-statement response} and introduces additional constructs not stated in the annotated hypothesis.
\end{minipage}
}
\caption{{Qualitative example of incorrect hypothesis extraction with high embedding similarity.}}
\label{fig:qual_hyp_example}
\end{figure*}

\begin{figure*}[htbp]
\centering
\setlength{\fboxsep}{6pt}
\setlength{\fboxrule}{0.6pt}
\fbox{
\begin{minipage}{0.97\linewidth}
\footnotesize
\textbf{Example: Hard-negative construction for within-document retriever training} \\

\textbf{DOI.} \href{https://doi.org/10.1016/j.jesp.2016.01.014}{10.1016/j.jesp.2016.01.014}\\

\textbf{Human annotation.}
\emph{We expected that participants experiencing causal uncertainty would show higher rates of task resumption when the interrupted task was associated with more abstract (vs. concrete) thinking.} \\

\textbf{Positive paragraph (contains human annotated hypothesis).}
\emph{\textbf{Returning to our main prediction, we expected that participants experiencing causal uncertainty would show higher rates of task resumption when the interrupted task was associated with more abstract (vs.\ concrete) thinking.} To measure task resumption, we gave participants an opportunity to engage in a more enjoyable task that involved reading humorous stories. The pilot study confirmed that the Humor Evaluation Task was indeed perceived as a more enjoyable alternative to participants. \ldots} \\

\textbf{Hard negative paragraph (topically similar, different rhetorical role).}
\emph{The task interruption and resumption paradigm used in this experiment rules out this alternative explanation because the alternative task that participants could choose to do involved reading humorous stories. The pilot study confirmed this, showing that participants thought the humor evaluation task would be the most fun and enjoyable. If participants' goal was to simply be in a better mood state, rather than to think more abstractly, we would not have found our demonstrated effect \ldots} 
\end{minipage}
}
\caption{{Example of the \emph{hard negative} selection for retriever training. Hard negatives are chosen to be semantically similar to the gold paragraph while differing in rhetorical function with respect to the extraction target.}}
\label{fig:hard_negative_example}

\end{figure*}

\begin{table*}[h]
\fontsize{8}{11}
\small
\selectfont{
\centering
\begin{tabular}{|l|l|lll|}
\hline
{LLM} & {Method} & \multicolumn{3}{c|}{{Evidence}}\\
& &  P & R & F1 \\
\hline
\multirow{6}{*}{llama 3.2} 
& RAG (k=5) &  0.19 & 0.25 & 0.22\\
& RAG (k=10) &  0.24 & 0.32 & 0.27\\
& RAG (k=20) &  0.25 & 0.29 & 0.26\\
& Full text  &  0.21 & 0.28 & 0.24 \\
\cline{2-5}
& RAG + Reranker (k=5) &  0.29 & 0.43 & 0.35\\
& Fine-tuned SciBERT + Reranker (k=5) &  0.31 & 0.48 & 0.38\\
& Oracle context &  0.43 & 0.52 & 0.47\\
\hline
\multirow{6}{*}{gpt-oss} 
& RAG (k=5) &  0.19 & 0.20 & 0.20 \\
& RAG (k=10) &  0.23 & 0.28 & 0.25\\
& RAG (k=20)  &  0.25 & 0.34 & 0.29 \\
& Full text  &  0.28 & 0.36 & 0.31 \\
\cline{2-5}
& RAG + Reranker (k=5) & 0.34 & 0.44 & 0.38\\ 
& Fine-tuned SciBERT + Reranker (k=5) &  0.43 & 0.49 & 0.46\\
& Oracle context &  0.50 & 0.57 & 0.53\\
\hline
\multirow{6}{*}{gpt-4o-mini} 
& RAG (k=5) &  0.22 & 0.40 & 0.28 \\
& RAG (k=10) &  0.24 & 0.45 & 0.31\\
& RAG (k=20)  &  0.28 & 0.49 & 0.36 \\
& Full text  &  0.35 & 0.40 & 0.37 \\
\cline{2-5}
& RAG + Reranker (k=5) &  0.39 & 0.45 & 0.41\\
& Fine-tuned SciBERT + Reranker (k=5) &  0.43 & 0.51 & 0.47\\
& Oracle context &  0.49 & 0.63 & 0.53\\
\hline
\multirow{6}{*}{gemini-2.5-flash}
& RAG (k=5) & 0.21 & 0.39 & 0.27\\
& RAG (k=10) & 0.24 & 0.44 & 0.31 \\
& RAG (k=20) & 0.29 & 0.50 & 0.37\\
& Full text  &  0.33 & 0.48 & 0.39 \\
\cline{2-5}
& RAG + Reranker (k=5) & 0.27 & 0.48 & 0.34\\
& Fine-tuned SciBERT + Reranker (k=5) & 0.31 & 0.53 & 0.38\\
& Oracle context & 0.51 & 0.59 & 0.53\\
\hline
\end{tabular}
\caption{Evidence extraction with ground-truth hypotheses. We isolate the evidence stage by providing the extractor the human-annotated (ground-truth) hypothesis and varying only the evidence-stage context configuration (RAG $k \in \{5,10,20\}$, reranking, and fine-tuned retrieval). Reported Precision/Recall/F1 are computed with the same component-based evaluator as in the main results.}
\label{table:c4_results}}
\end{table*}

\end{document}